\documentclass[runningheads]{llncs}
\usepackage{graphicx}

\usepackage{tikz}
\usepackage{comment}
\usepackage{amsmath,amssymb} 
\usepackage{color}
\usepackage{graphicx}
\usepackage{booktabs}
\usepackage{overpic}
\usepackage{soul}
\usepackage{float}
\usepackage{adjustbox}
\usepackage{wrapfig}
\usepackage{abstract}
\usepackage{xcolor}
\usepackage{subcaption}
\usepackage{tabularx}
\usepackage{array}
\newif\ifdraft
\drafttrue

\ifdraft
\newcommand{\sagie}[1]{{\color{red}[\textbf{Sagie:} #1]}}
\newcommand{\sebastian}[1]{{\color{green}[\textbf{Sebastian:} #1]}}

\else
\newcommand{\sagie}[1]{}
\newcommand{\sebastian}[1]{}
\fi

\newif\ifreviewer
\reviewertrue
\reviewerfalse
\ifreviewer
\newcommand{\rev}[1]{{\color{black}#1}}
\else
\newcommand{\rev}[1]{}
\fi

\usepackage[colorlinks,citecolor=green,urlcolor=blue,bookmarks=false,hypertexnames=true]{hyperref}

\usepackage[capitalize]{cleveref}
\crefname{section}{Sec.}{Secs.}
\Crefname{section}{Section}{Sections}
\Crefname{table}{Table}{Tables}
\crefname{table}{Tab.}{Tabs.}

\usepackage[accsupp]{axessibility}  

\usepackage[width=122mm,left=12mm,paperwidth=146mm,height=193mm,top=12mm,paperheight=217mm]{geometry}
\usepackage{hyperref}

\begin{document}
\pagestyle{headings}
\mainmatter
\def\ECCVSubNumber{*****}  

\title{Text-Driven Stylization of Video Objects} 

\title{Text-Driven Stylization of Video Objects} 
\author{Sebastian Loeschcke$^1$, Serge Belongie$^2$, Sagie Benaim$^2$}
\institute{$^1$Aarhus University, $^2$University of Copenhagen}

\titlerunning{Text-Driven Stylization of Video Objects}
\authorrunning{Loeschcke et al., }

\maketitle

\begin{abstract}

We tackle the task of stylizing video objects in an intuitive and semantic manner following a user-specified text prompt. 
This is a challenging task as the resulting video must satisfy multiple properties: (1) it has to be temporally consistent and avoid jittering or similar artifacts, (2) the resulting stylization must preserve both the global semantics of the object and its fine-grained details, and (3) it must adhere to the user-specified text prompt. To this end, our method stylizes an object in a video according to two target texts. The first target text prompt describes the global semantics and the second target text prompt describes the local semantics. To modify the style of an object, we harness the representational power of CLIP to get a similarity score between (1) the local target text and a set of local stylized views, and (2) a global target text and a set of stylized global views. We use a pretrained atlas decomposition network to propagate the edits in a temporally consistent manner. We demonstrate that our method can generate consistent style changes over time for a variety of objects and videos, that adhere to the specification of the target texts. We also show how varying the specificity of the target texts and augmenting the texts with a set of prefixes results in stylizations with different levels of detail. 
Full results are given on our project webpage:  \url{https://sloeschcke.github.io/Text-Driven-Stylization-of-Video-Objects/}. 


\end{abstract}

\section{Introduction}

Manipulating semantic object entities in videos using human instructions requires skilled workers with domain knowledge.
We seek to eliminate these requirements by specifying a desired edit or stylization through an easy, intuitive, and semantic user instruction in the form of a text-prompt. 

However, manipulating video content semantically is a challenging task. One challenge is in generating consistent content or style changes in time, that adhere to the target text specification. Another challenge is to manipulate the content of an object such that it preserves the content of the original video and the global semantics while also adhering to fine-grained details in the target text.  

\begin{figure*}[t]
\centering
\begin{tabular}{ccc}

         \includegraphics[width=0.32\linewidth]{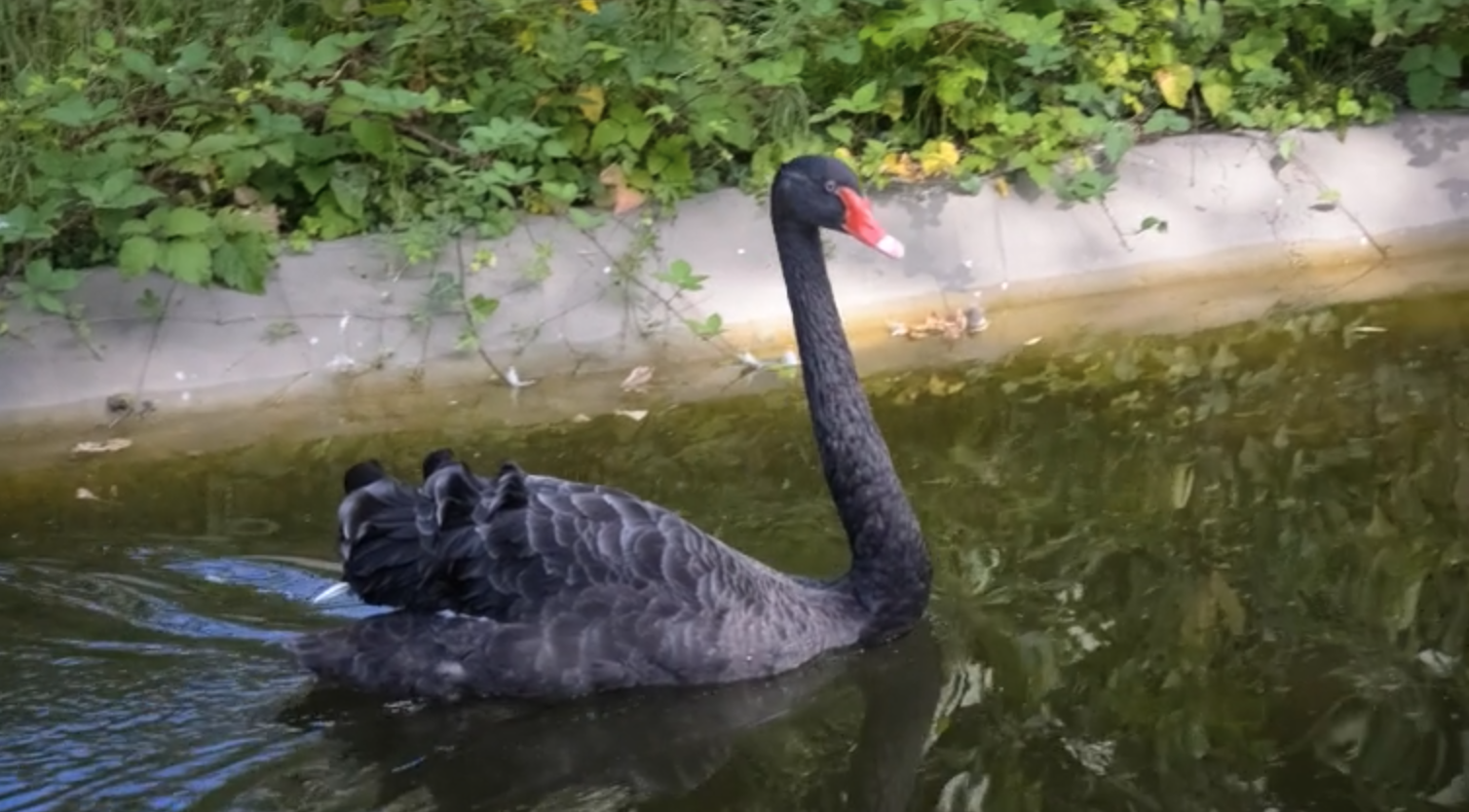} &
         \includegraphics[width=0.32\linewidth]{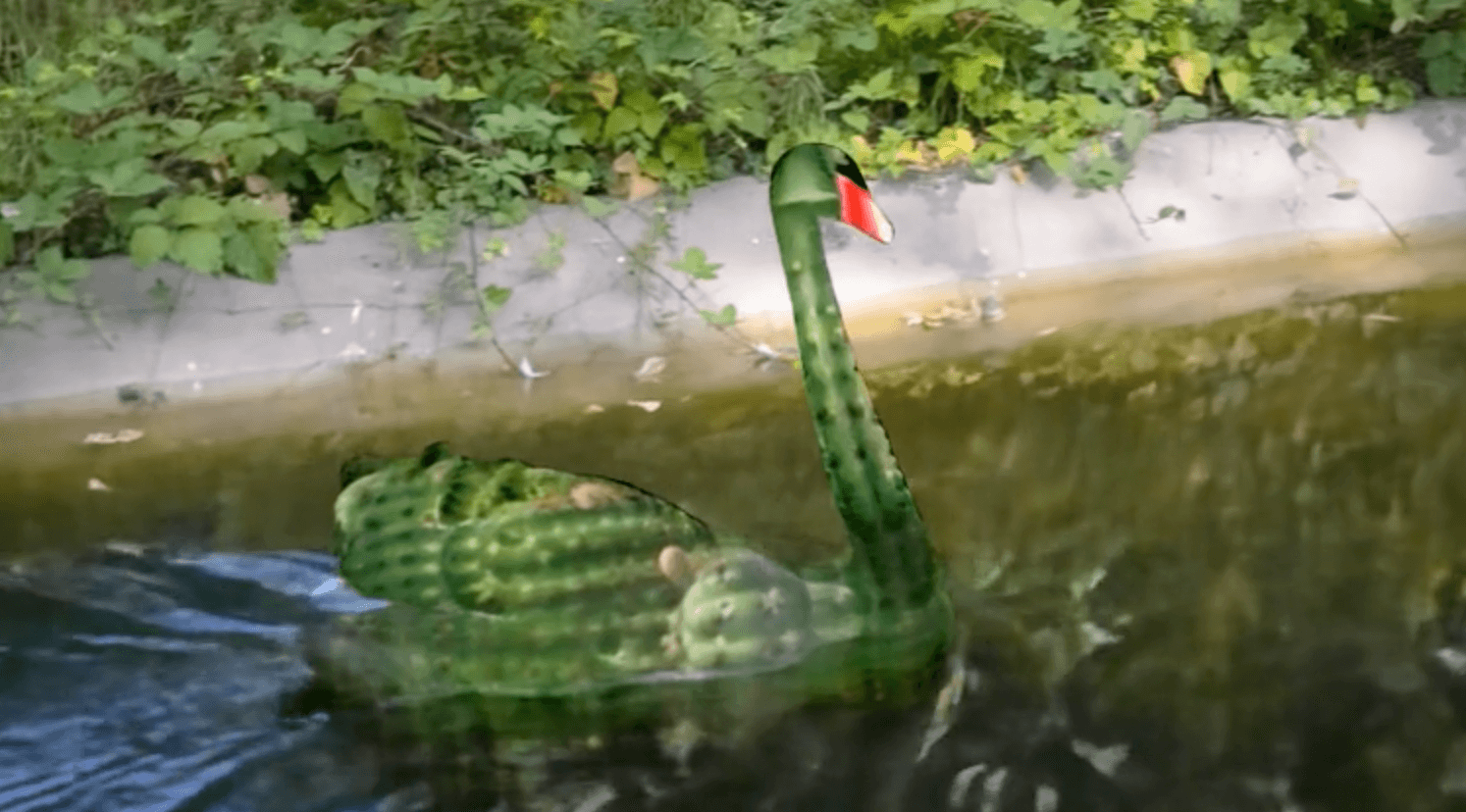} &
         \includegraphics[width=0.32\linewidth]{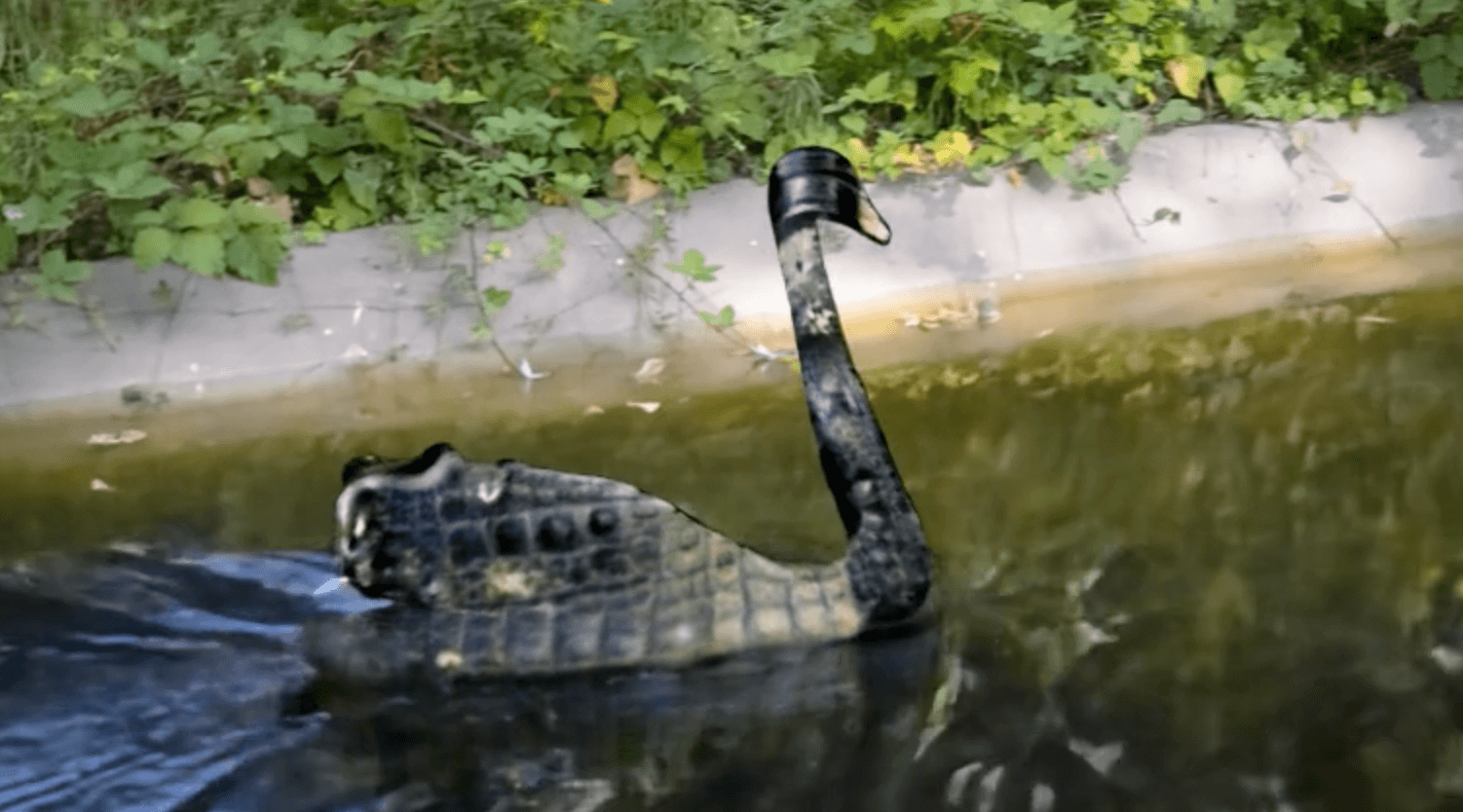} \\
         
         \includegraphics[width=0.32\linewidth]{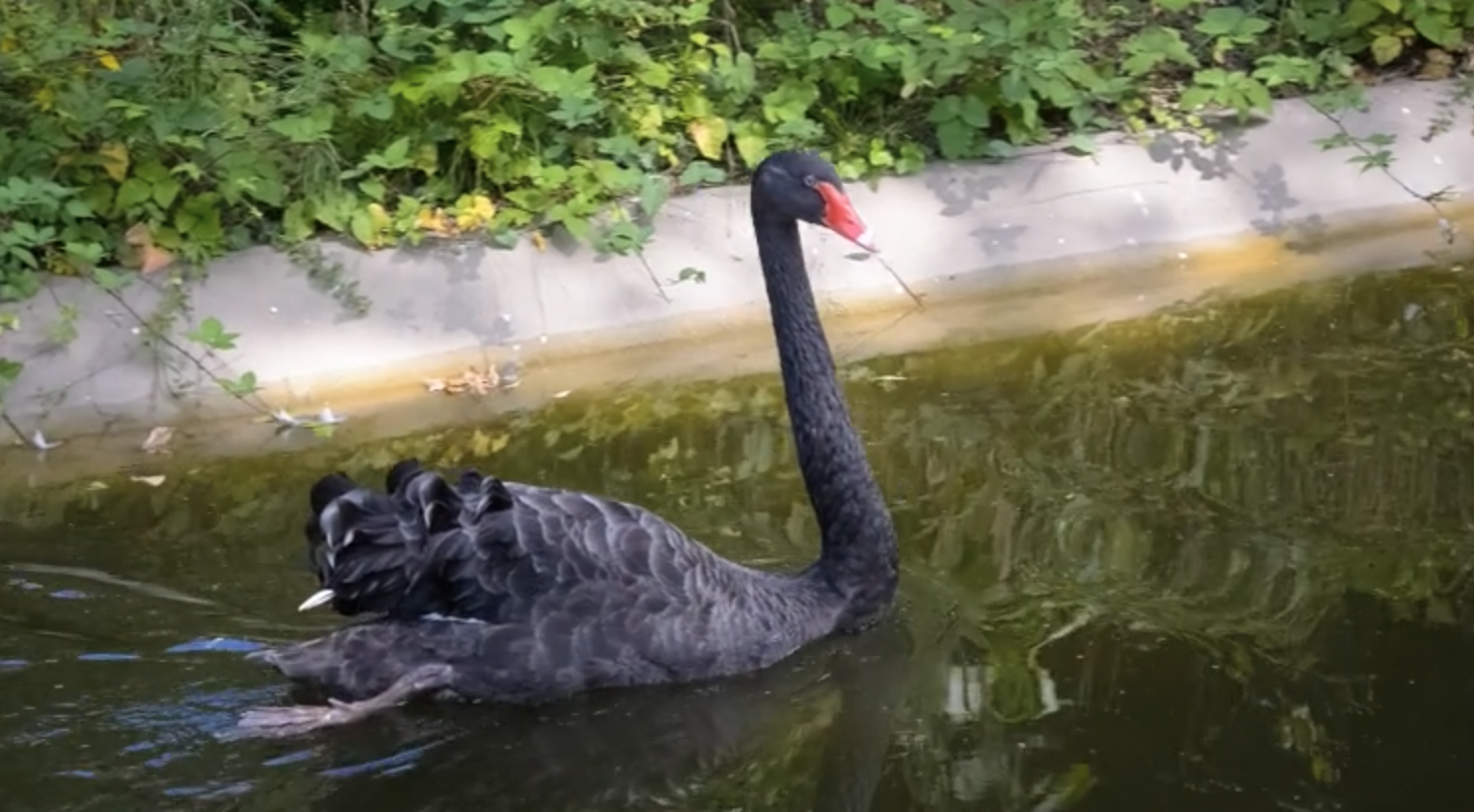} &
         \includegraphics[width=0.32\linewidth]{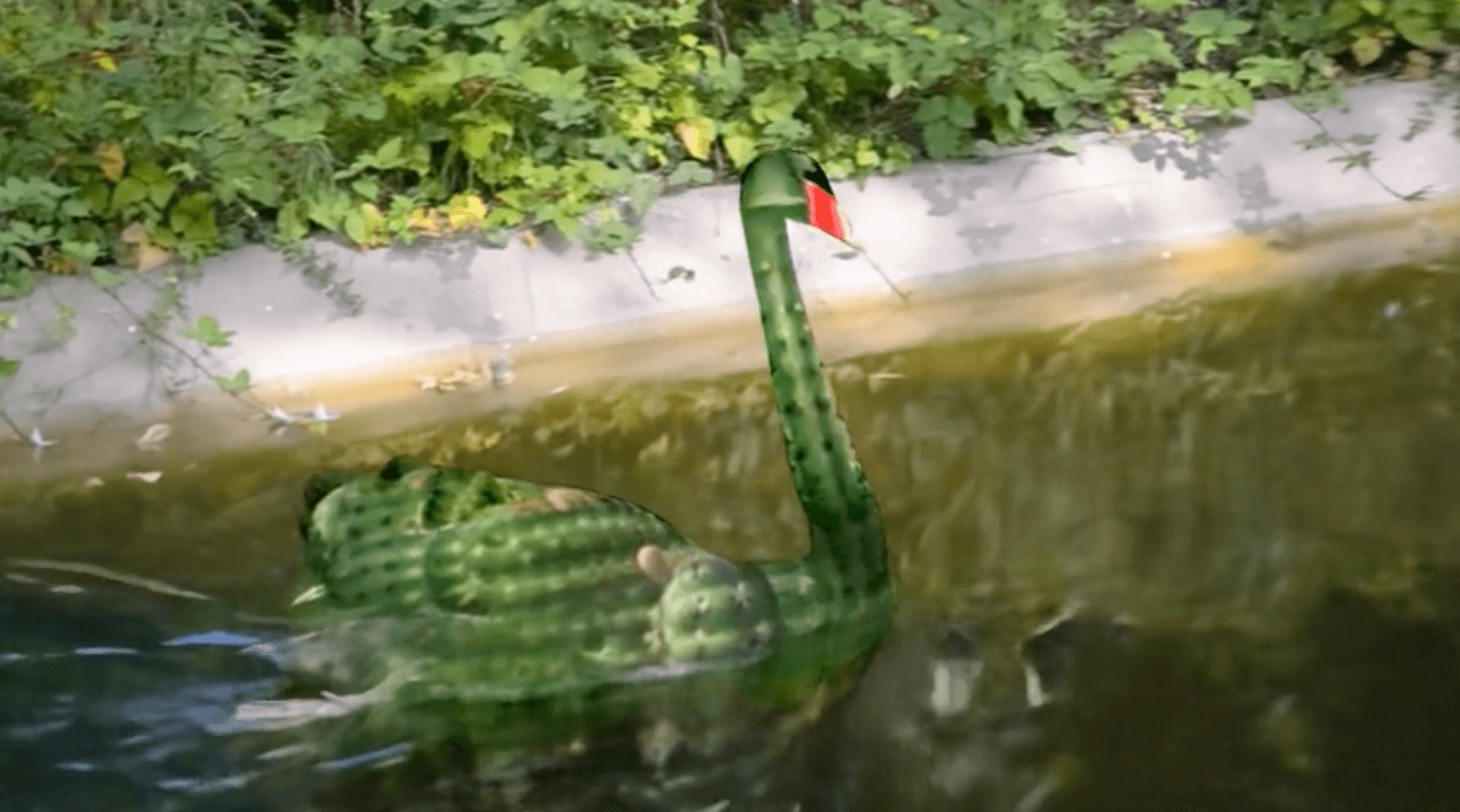} &
         \includegraphics[width=0.32\linewidth]{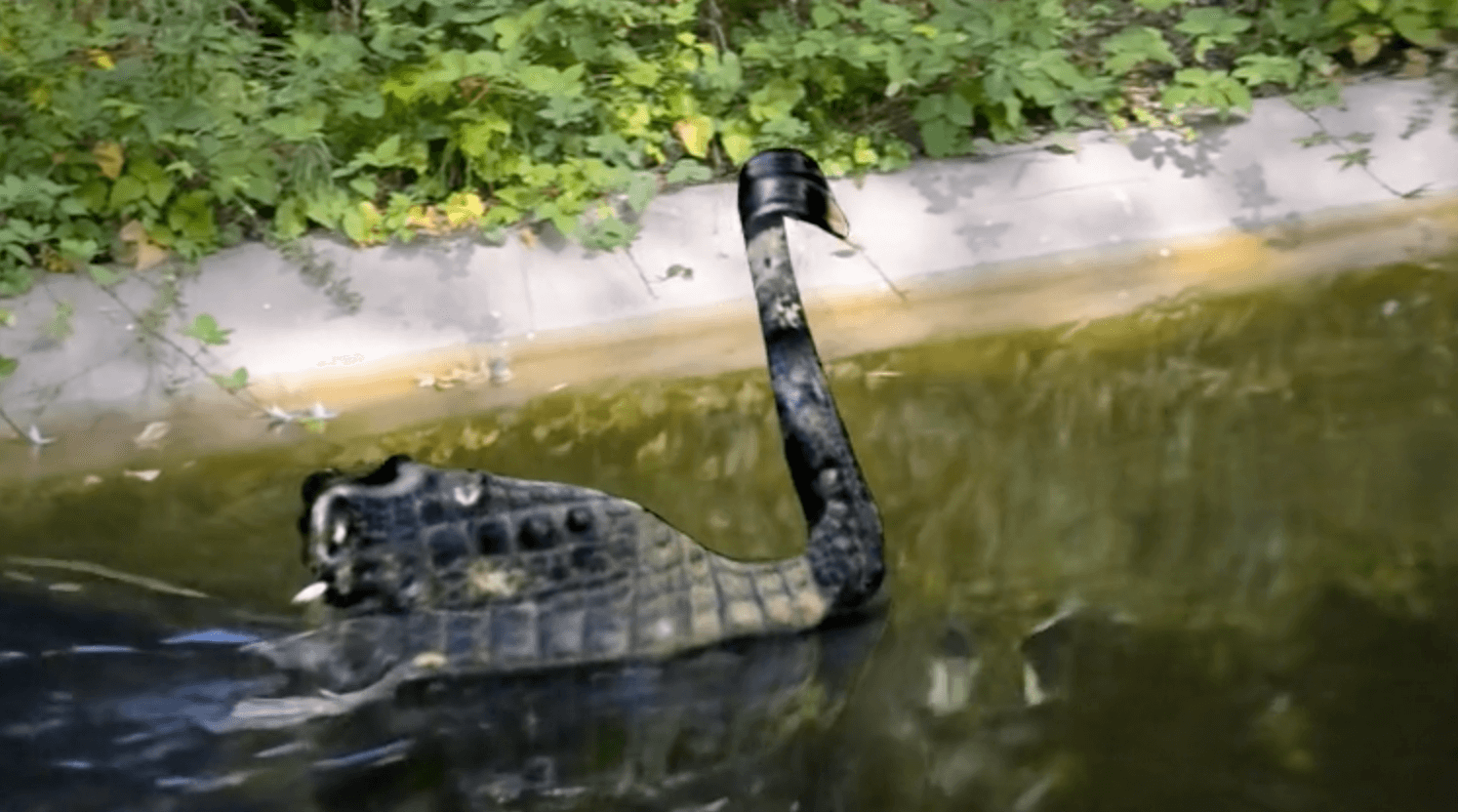} \\
         Input video &``Swan made out of cactus'' &``Swan with crocodile skin'' \\
         \vspace{-0.5cm}
\end{tabular}
     \caption{Two representative video frames and the edited video frames together with the global target text.} 
     \vspace{-0.5cm}
    \label{fig:teaser}
\end{figure*}

In recent years, advances in computational methods emerged that enable manipulation of appearances and style in images and allow novice users to perform realistic image editing. These methods include manipulation tools that use natural language (text prompts) to express the desired stylization of images or 3D objects~\cite{Text2Mesh,VQGAN-CLIP}. The text-driven manipulation is facilitated by recent developments in models for joint embeddings of text and images, e.g the Contrastive Language Image Pretraining (CLIP~\cite{CLIP}) model. Instead of manipulating images or 3D objects, we use CLIP in the context of video manipulation. 
This is not a straightforward task since simply maximizing the semantic (CLIP-based) similarity between a valid target text and each 2D frame in the video often leads to degenerate solutions. Also, applying methods for image manipulation to each frame in a video results in edits that lack temporal consistency. 

The recently introduced Neural Layered Atlases (NLA) work~\cite{LayeredNeuralAtlases}, demonstrates the ability to separate a moving object in a video from its background by decomposing the video into a set of 2D atlases. Each atlas provides a unified image representation of an object or background over the video. Edits applied to the image representation are automatically mapped back to the video in a temporally consistent manner. However, editing an image still requires manual effort and editing skills from the user. Another problem with this approach is that the 2D atlas representation can be hard to edit due to local deformations.

We propose a method for performing intuitive and consistent video editing with multiple capabilities by using the representational power of CLIP to express a desired edit through a text-prompt. An example could be to change the style of a swan swimming in a lake according to a target text: ``A swan with cactus skin.'' Text is easily modifiable and allows users to express complex and abstract stylizations intuitively. Using text to express edits reduces the need for manual editing skills and also avoids the problems related to deformation in the 2D atlas representations. An illustration is shown in \cref{fig:teaser}. 

To apply temporally consistent edits to an object in a video, our method uses the atlas decomposition method presented in NLA~\cite{LayeredNeuralAtlases}. We train a generator on a single input video by sampling local and global views of each frame in the video and applying various augmentations to each view. Our method uses a global loss that compares each global view with a global target text and a local loss that compares each local view with a local target text. The global loss then focuses on the global semantics and the local views focus on the fine-grained details. To regularize our learning, we use a sparsity loss that encourages sparse representation and a temporal triplet loss that encourages frames that are close in time to be similar in CLIP's embedding space. 

We demonstrate that our method results in natural and consistent stylizations of objects for a diverse set of videos and target texts. We show how varying the specificity of both the local and global target texts varies the stylization and how augmenting the target texts with neutral prefixes can result in more detailed stylizations. We also demonstrate that our global loss focuses on the global semantics and the local losses on the fine-grained details.

\section{Related Work}
Our work is related to video editing works and also to text-based stylization works. 

\subsection{Video editing}
Unlike images, editing or stylizing objects in videos requires the ability to handle temporal consistency. 
One natural approach is to propagate the edits from one frame to the next. Video Propagation Networks~\cite{VideoPropagationNetworks}  use a bilateral network to connect pixels in consecutive frames and an adaption network to refine the pixels. Other approaches use optical flow to propagate edits made on a few key-frames~\cite{key_frame_edits}. These approaches work well when there is a clear correspondence between frames, but have difficulties, e.g., when the video contains occlusions. 

To address occlusion  challenges, recent work has used deep learning approaches, e.g. self-supervised methods for learning visual correspondence from unlabeled videos~\cite{VideoLablProb,VideoLablProb2}.
Instead, our work uses the representation proposed by Neural Layered Atlases (NLA)~\cite{LayeredNeuralAtlases}, which decomposes a video into a set of layered 2D atlases. Each atlas provides a unified representation of the appearance of an object or background throughout the video. Similar to NLA, Deformable Sprites~\cite{DeformableSprites} decomposes a video into a texture atlas that captures an object's motion across the entire video. Their method allows for temporally consistent video edits, by applying edits to the decomposed atlas. In contrast to NLA, they use optical flow to compute foreground objects instead of a pretrained segmentation network to get the segmentation mask for an object in the input video. 
However, both NLA and Deformable Sprites only allow for basic manual editing. We use NLA's atlas separation method for objects in videos, but unlike NLA, we allow for text-driven stylization.

\subsection{Text-based stylization}
Our work bears similarities to recent image and 3D manipulation techniques that edit style and appearances through natural language descriptions. These descriptions are often embedded with the Contrastive Language Image Pretraining (CLIP)~\cite{CLIP} model, a multi-modal embedding model that learns an image-text embedding space. Recent work used CLIP together with pretrained generative networks for image editing and stylization~\cite{blended_diffusion,StyleCLIP,PaintByWord,Image-stylization,VQGAN-CLIP,StyleGAN-nada,DiffusionCLIP}. For example, StyleCLIP~\cite{StyleCLIP}, and StyleGAN-NADA~\cite{StyleGAN-nada} both use a pretrained StyleGAN~\cite{StyleGAN} and CLIP to perform image editing, either by using CLIP to control a latent code or to adapt an image generator to a specific domain~\cite{StyleGAN-nada}. Other examples include using StyleGAN and CLIP for image stylization~\cite{Image-stylization} or Paint by Word~\cite{PaintByWord} which uses CLIP paired with StyleGAN2~\cite{StyleGAN2} and BigGAN~\cite{BigGan} to enable ``painting'' images in the style of a text prompt. 
Usually, pretrained generators only work well for the specific domain they are trained on. In contrast, our method does not require a pretrained generator. We train our own generator on the set of video frames we wish to stylize. 

Other examples of semantic text-guided manipulation in the context of 3D objects include 3DStyleNet~\cite{3DStyleNet}, a method for changing the geometric and texture style of 3D objects, ClipMatrix~\cite{ClipMatrix} which uses text-prompts to create digital 3D creatures, and methods for generating 3D voxels using CLIP~\cite{3dVoxel}. 
Another line of recent work uses joint-embedding architectures~\cite{imagen_google_brain,Dall-e,dall-e2,GLIDE} for image generation, e.g., DALL-E~\cite{Dall-e} and its successor, DALL-E 2~\cite{dall-e2}, which can also be used for stylizing images. DALL-E 2 uses a two-stage model, where a CLIP image embedding is generated using a text prompt and a decoder is then used to generate an image conditioned on the generated image embedding.
Training joint embedding architectures requires enormous datasets and many training hours. Instead of training on a large dataset, we train on a set of frames for a single video and use augmentations to extract many different views of the input frames.
As opposed to all the abovementioned techniques, we work on videos.

Another line of work uses CLIP without relying on a pretrained generator, e.g. given a natural language input, CLIPdraw~\cite{CLIP_Draw} synthesizes novel drawings, and CLIPstyler~\cite{CLIPstyler} stylizes images. Similarly, Texts2Mesh~\cite{Text2Mesh} does not rely on a pretrained generator. In Text2Mesh, the CLIP embedding space is used to enable text-driven editing of 3D meshes. Text2Mesh uses a multi-layer perceptron (MLP) to apply a stylization to $(x,y,z)$-coordinates of an input mesh. The neural optimization process of the MLP is guided by a semantic loss that computes the similarity between multiple augmented 2D views embedded with CLIP and a target text. Similarly, we do not rely on a pretrained generator. 

Our work was developed concurrently to Text2Live~\cite{Text2Live}, which shares many of the same goals and methods as our work. Similarly to our method, Text2Live uses a pretrained Neural Layered Atlases (NLA) model to separate a moving object in a video from its background. Text2Live train a generator to apply text-driven local edits to a single frame and use the NLA model to map the edits back to the input video in a temporally consistent manner. In contrast to our approach, Text2Live does not directly generate the edited output. Instead, it generates an edit layer that is composited with the original input. 

\section{Method}
We wish to apply natural and temporally consistent stylizations to objects in videos using a natural language text prompt as guidance. To change the style of an object to conform with a target text prompt in a temporally consistent manner, we build on top of a Layered Neural Atlas method~\cite{LayeredNeuralAtlases}, which separates the appearance of an object in a video from its background. We then use a pre-trained text-image multimodal embedding of CLIP~\cite{CLIP} in a set of objectives. Minimizing this set of objectives aims at matching the style of a foreground object in a video with that of a target text. 
The objectives include a global and local objective. The global objective focuses on the global semantics by maximizing the similarity between the global views and a target text that relates to the underlying content. Instead, the local objective focuses on the fine-grained details, by maximizing the similarity between the local views with a target text that relates to local semantics of the stylization. 
We add a sparsity loss, similar to a $L_1$-regularization term, that encourages the predicted foreground color values to be minimal. Additionally, we add a temporal triplet loss that encourages the embeddings of frames that are close in time to also be close in CLIP's embedding space.  
We begin by describing the method of CLIP~\cite{CLIP} and that of Neural Layered Atlases (NLA) on which our method is based. We then describe the training and loss formulations used by our method. 

\subsection{CLIP}
CLIP is a multi-modal embedding method that trains an image encoder $E_{img}$ and a text encoder $E_{txt}$ to match between the embeddings of corresponding image-text pairs using a contrastive loss formulation. 
This loss formulation optimizes the similarity between corresponding image-text pair $T$ and $I$. 
More specifically, $I$ and $T$ are first embedded: $$ I_{emb} = E_{img}(I) \in \mathbb{R}^{512}, \quad T_{emb} = E_{txt}(T) \in \mathbb{R}^{512}$$
The similarity between I and T is then measured by 
 $\operatorname{sim}(I_{emb}, T_{emb} )$ 
 where $\operatorname{sim}(a, b)=\frac{a \cdot b}{|a||b|}$, is the cosine similarity.

\subsection{Neural Layered Atlases (NLA)}

\begin{figure}
    \centering
    \includegraphics[width=1.0\textwidth]{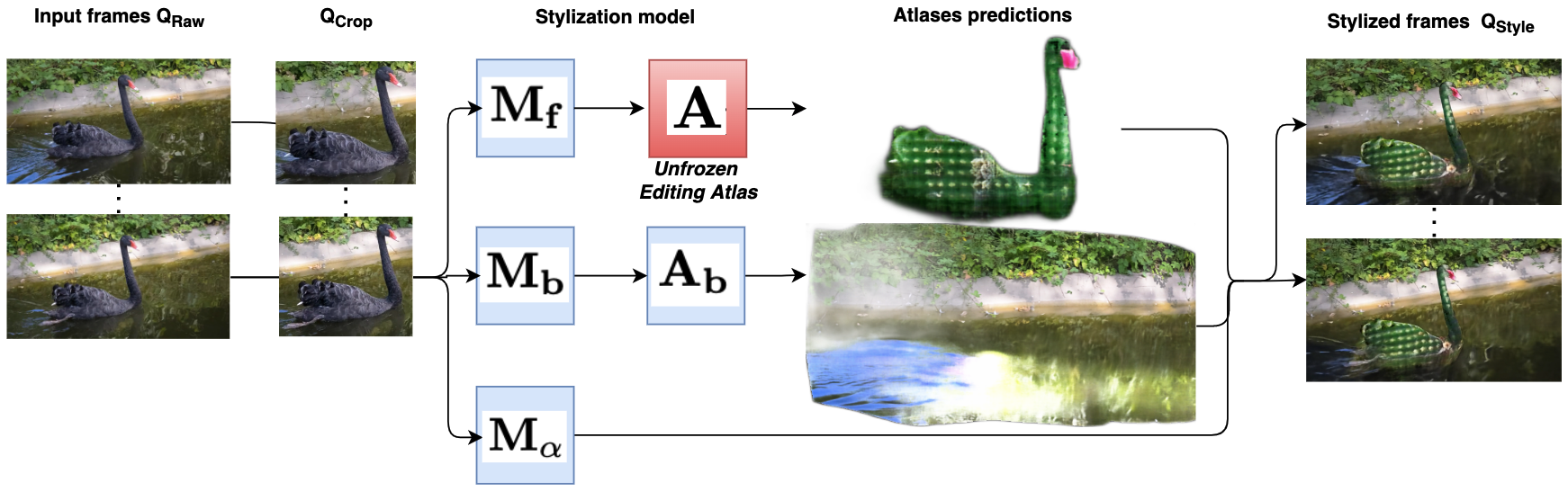}
    \caption{\textbf{Our stylization pipeline.} In the first step, we train the network using the NLA procedure~\cite{LayeredNeuralAtlases} to reconstruct input video frames. We then finetune the editing atlas $A$ using our approach described in Fig.~\ref{fig:method_diagram}. In our stylization pipeline, we create a set of cropped input video frames $Q_{Crop}$. This set is passed through a stylization model to create a foreground and background atlas. A set of stylized frames $Q_{Style}$ is produced by $\alpha$-blending the predicted atlases. All weights of the MLPs are \textcolor{blue}{frozen} except for the editing atlas MLP $A$ which is \textcolor{red}{finetuned}. A closer look at how our editing atlas is trained is given in \cref{fig:method_diagram}.}
    \label{fig:NLA_method}
\end{figure}

\begin{figure}
    \centering
    \includegraphics[width = \textwidth]{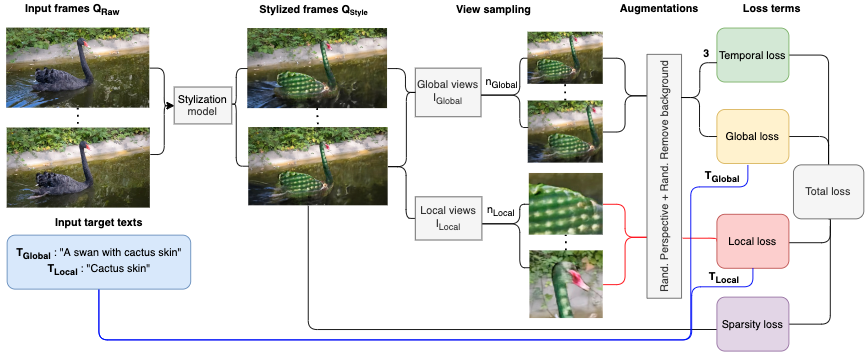}
    \caption{\textbf{Finetuning the editing atlas.} 
    As described in the stylization pipeline (Fig.~\ref{fig:NLA_method}), we use our stylization pipeline to get a set of stylized frames $Q_{Style}$. To this end, we finetune our editing atlas, which is part of the stylization model. We sample $n_{Global}$ global views $I^{Global}$ and $n_{Local}$ local views $I^{Local}$. The set of images $I^{Global}$ are then augmented using random perspectives and random background removal and used together with the \textcolor{blue}{global target text $T_{Global}$} to compute a \textcolor{yellow}{global loss} (Eq.~\ref{global_loss_function}). Three global augmented images are used to compute the  \textcolor{green}{temporal loss} (Eq.~\ref{temporal_loss_function}). Similarly, the $I^{Local}$ images are augmented and used together with the  \textcolor{blue}{local target text $T_{Local}$} to compute the \textcolor{red}{local loss} (Eq.~\ref{local_loss_function}). Lastly, a \textcolor{purple}{sparsity loss} (Eq.~\ref{sparsity_loss_function}) is computed from the stylized frames $Q_{Style}$.}
    \label{fig:method_diagram}
\end{figure}

 Neural Layered Atlases (NLA)~\cite{LayeredNeuralAtlases} decompose a video into a set of layered 2D atlases. Each atlas provides a unified representation of the appearance of an object or the background throughout the video. NLA use two mapping networks $M_f$ and $M_b$, where each takes a pixel and time  location $(x,y,t)$ in the video as input and outputs the corresponding 2D $(u,v)$-coordinate in each atlas: 
$$
 M_f(p)=(u_f^p,u_f^p), \quad M_b(p)=(u_b^p,u_b^p)
$$ 
An atlas network $A$ takes the predicted 2D $(u,v)$-coordinate as input and outputs the atlas’s RGB color at that location. Additionally, all pixel coordinates are fed into the Alpha MLP network $M_\alpha$ which outputs the opacity of each atlas at that location. The RGB color can then be reconstructed at each pixel location by alpha-blending the predicted atlas points according to the opacity value predicted by $M_\alpha$.

NLA enables consistent video editing. First, each atlas is discretized into an image. A user can then manually apply edits using an editing program. These edits are mapped back to the input video using the computed $(u,v)$-mapping. To get the reconstructed color $c^p$ for pixel $p$, the color of the predicted foreground color $c_f^p$, background color $c_b^p$ and predicted opacity value $\alpha^p$ of the edited atlas are blended as follows: 
$$c^p=(1-\alpha^p)c_b^p+\alpha^pc_f^p$$

\subsection{Our Stylization Pipeline}
Instead of having to manually apply edits to the discretized atlases as in~\cite{LayeredNeuralAtlases}, we wish to apply edits automatically using a target text prompt. Specifically, we are interested in modifying the RGB values such that they conform with a target text. We focus on the Atlas MLP $A$, since $A$ makes the RGB predictions. Instead of using a single atlas MLP $A$ for all atlases, we create two copies of the atlas MLP after pre-training the NLA network, one editing atlas MLP for the foreground object we want to stylize ($A$) and one for all other atlases ($A_b$). We then freeze the weights of $M_b$, $M_f$, $M_\alpha$, $A_b$ and finetune the weights of $A$. While the goal in NLA is to optimize the model to produce the best reconstruction, we instead want to create a stylization that conforms with the inputted target text. 

 Our stylization pipeline is illustrated in Fig.~\ref{fig:NLA_method}. The input is a set of raw frames $Q_{Raw}$. Since we know the position of the object in the video from the $\alpha$-map, we can compute a bounding box containing the whole object. Once we have the bounding box, we crop each frame in $Q_{Raw}$ such that it only contains the content within the bounding box plus a small margin.
All the cropped frames $Q_{Cropped}$ are passed through a pre-trained NLA model, where all MLP weights have been frozen, except for the weights of the $A$ MLP. The NLA method produces a set of stylized frames $Q_{Style}$.

To fine-tune the weights of $A$, we sample training batches in both time and space. Our sampling method is illustrated in Fig.~\ref{fig:method_diagram}. First, we sample a set $Q_{Sample}$ uniformly at random among all frames in the input video and pass them through the stylization pipeline (Fig.~\ref{fig:NLA_method}) to create a set $Q_{Style}$. 

For each frame in $Q_{Style}$ we sample $n_{Global}$ views $I^{Global}$ and a set of $n_{Local}$ views $I^{Local}$. 
Each of the $n_{Global}$ views is produced by sampling a crop with a size in the range $[0.9,1.0]$ of the original frame size. Each of the $n_{Local}$ views is produced by sampling a crop with a size in the range $[0.1,0.5]$ of the original frame size. To ensure the local views contain the object we want to stylize, we use the $\alpha$-map of the frame to determine the position of the object in the frame. 
We then sample until we get $n_{Local}$ views where at least $\frac{1}{3}$ of the sampled view is part of the object. 

Once the local and global views have been sampled, we apply a random perspective transformation and a random background removal which with some probability $p$ removes the background (details in Appendix \cref{sec:appendix_img_augmentations}). 
Additionally, each augmented frame is normalized with the same mean and standard deviation as used to train the CLIP model~\cite{CLIP}.

Our objective function is composed of three main losses defined in CLIP's feature space: (1) $L_{\text {Local}}$ which focuses on local semantics, (2) $L_{\text {Global }}$ which focuses on global semantics, and
(3) $L_{T e m p}$ which encourages temporal consistency. Additionally, we use the regularization term $L_{\text {sparsity }}$ introduced in NLA~\cite{LayeredNeuralAtlases}, which encourages sparse representations. 
In all loss terms, when we compute the similarity between a text and each sampled view, we use the average embedding across all views:
\begin{equation} \label{avg_emb}
I_{emb}^{Local} = \frac{1}{n_{Local}} \sum_{i=1}^{n_{Local}}E_{i m g}\left(I_i^{\text {Local}}\right), I_{emb}^{Global} = \frac{1}{n_{Global}} \sum_{i=1}^{n_{Global}}E_{i m g}\left(I_i^{\text {Global}}\right) 
\end{equation}

\paragraph{Local loss}
$L_{\text {Local}}$ is applied to all views in $I^{\text {Local}}$. The goal is to modify the image, such that the local details conform with a target text $T_{\text {Local}}$:
\begin{equation}
\label{local_loss_function}
L_{\text {Local}}= 1 - \operatorname{sim}\left(I_{emb}^{Local} , E_{t x t}\left(T_{\text {Local}}\right)\right)
\end{equation}
where $\operatorname{sim}(a, b)=\frac{a \cdot b}{|a||b|}$ is the cosine similarity, and $E_{t x t}$ denote CLIP's pre-trained text encoder.
The local views have a more zoomed-in view of the object we are stylizing. Additionally, the local target text $T_{Local}$ contains local specific semantics, e.g. ``\textit{rough} cactus texture.'' Hereby, the local loss can focus on the texture and fine-grained details of the stylization we apply to the input video.

\paragraph{Global loss}

The global loss is applied to views in $I^{\text {Global}}$ that all include the entire object being stylized. The intended goal is that the global loss will preserve the overall context. In the target text $T_{\text {Global}}$, we include words that describe the global context of the object we are trying to stylize, e.g. ``A \textit{swan} made of cactus.'' The global loss formulation is then given by:
\begin{equation}
\label{global_loss_function}
L_{\text {Global}}= 1 - \operatorname{sim}\left(I_{emb}^{Global} , E_{t x t}\left(T_{\text {Global }}\right)\right)
\end{equation}

\paragraph{Temporal loss}
We use a triplet loss to include a temporal aspect and enforce that consecutive frames should be more similar in CLIP's embedding space than frames that are further apart. To compute the temporal loss, we sample three frames $t_{1}, t_{2,}, t_{3}$, where we have that $t_{1}<t_{2,}<t_{3}$ w.r.t. the order of the frames in the input video. We then enforce that the similarity between the sampled global views of $t_{1}$ and $t_{2}$ in the CLIP embedding space should be greater than the similarity between $t_{1}$ and $t_{3}$: 
Let $I_{emb({t_1})}^{Global}, \space{ }I_{emb({t_2})}^{Global},\space I_{emb({t_3})}^{Global}$ denote the average embedded global views (computed in Eq.~\ref{avg_emb}) for each of the three frames. Then we compute the triplet loss $L_{Temp}$ as follows:
\begin{equation} \label{temporal_loss_function}
\begin{split}
\operatorname{Sim}_{t_1 t_3-t_1 t_2} &=\operatorname{sim} \left(I_{emb({t_1})}^{Global}, I_{emb({t_3})}^{Global}\right)-\operatorname{sim}\left(I_{emb({t_1})}^{Global}, I_{emb({t_2})}^{Global}\right) \\ 
L_{T e m p} &= \lambda_{Temp} \cdot \max \left(0, \operatorname{Sim}_{t_1 t_3-t_1 t_2}\right) 
\end{split}
\end{equation}
where $\lambda_{Temp}$ is a weighting of the temporal loss. If the frames are further away from each other, the temporal loss should contribute less to the overall loss. For this reason, we weigh the contribution of the temporal loss by a Gaussian probability density function $g$ with a mean equal to zero and a standard deviation of five. We compute the weight of a triplet by applying $g$ to the difference between $t_1$ and $t_3$:
$$\lambda_{Temp}=g(t_3-t_1)$$

\paragraph{Sparsity loss}
We use the same sparsity loss as in NLA~\cite{LayeredNeuralAtlases}. Its intended function is to encourage points that are mapped to the background atlas to have a zero value in the foreground atlas, e.g. if a point $p$ is mapped to the background atlases it should not contain information about the foreground atlas.
\begin{equation} \label{sparsity_loss_function}
L_{\text {sparsity }}=\left\|\left(1-\alpha^{P}\right) c_{f}^{P}\right\|
\end{equation}
where $c_{f}^{P}$ is the predicted color at $p$ for the foreground layer and $\alpha^{P}$ is the opacity value at location $p$.

\paragraph{Full objective}
The full loss term that is minimized is represented as:
\begin{equation}
\label{objective function}
L=\lambda_{Sparsity}L_{\text {Sparsity }} + \lambda_{Temp}L_{\text {Temp }} + \lambda_{Local}L_{\text {Local}}+\lambda_{Global}L_{\text {Global }}
\end{equation}
where $\lambda_{Local}$,$\lambda_{Global}$, $\lambda_{Temp}$, $\lambda_{Sparsity}$ are hyperparameters used to control the weighting of each loss term. As default $\lambda_{Local}=\lambda_{Global}=1$, while $\lambda_{Temp}$ and and $\lambda_{Sparsity}$ vary depending on the input video.

\section{Experiments}

We evaluate our method on a set of videos from the DAVIS dataset~\cite{DAVIS_dataset} across a diverse set of target text prompts. Our goal is to perform consistent and natural video editing. For this purpose, we present both a quantitative and qualitative evaluation of our results and perform a careful ablation study of each loss term in our objective function.

In Sec.~\ref{section:general_results} we demonstrate the capabilities of our method by showing various stylizations for different videos.
We also present a quantitative evaluation, where we compare our method to an image baseline method applied to each frame in the video.
In Sec.~\ref{section:prompt_specificity} we demonstrate that the specificity of text prompts influences the details of the results. 
In Sec.~\ref{section:text_aug} we demonstrate how text augmentation affects the results of the stylizations.
In Sec.~\ref{section:ablation_study} we conduct a series of ablations on our loss terms and demonstrate how the local and global losses focus on different semantics. 
Finally, in Sec.~\ref{section:limitations} we illustrate some of the limitations of our method.

\label{sec:results}
\begin{figure*}
\centering
\begin{tabular}{ccc}
         \includegraphics[width=0.32\linewidth]{images/example_images/org_swan1.png} &
         \includegraphics[width=0.32\linewidth]{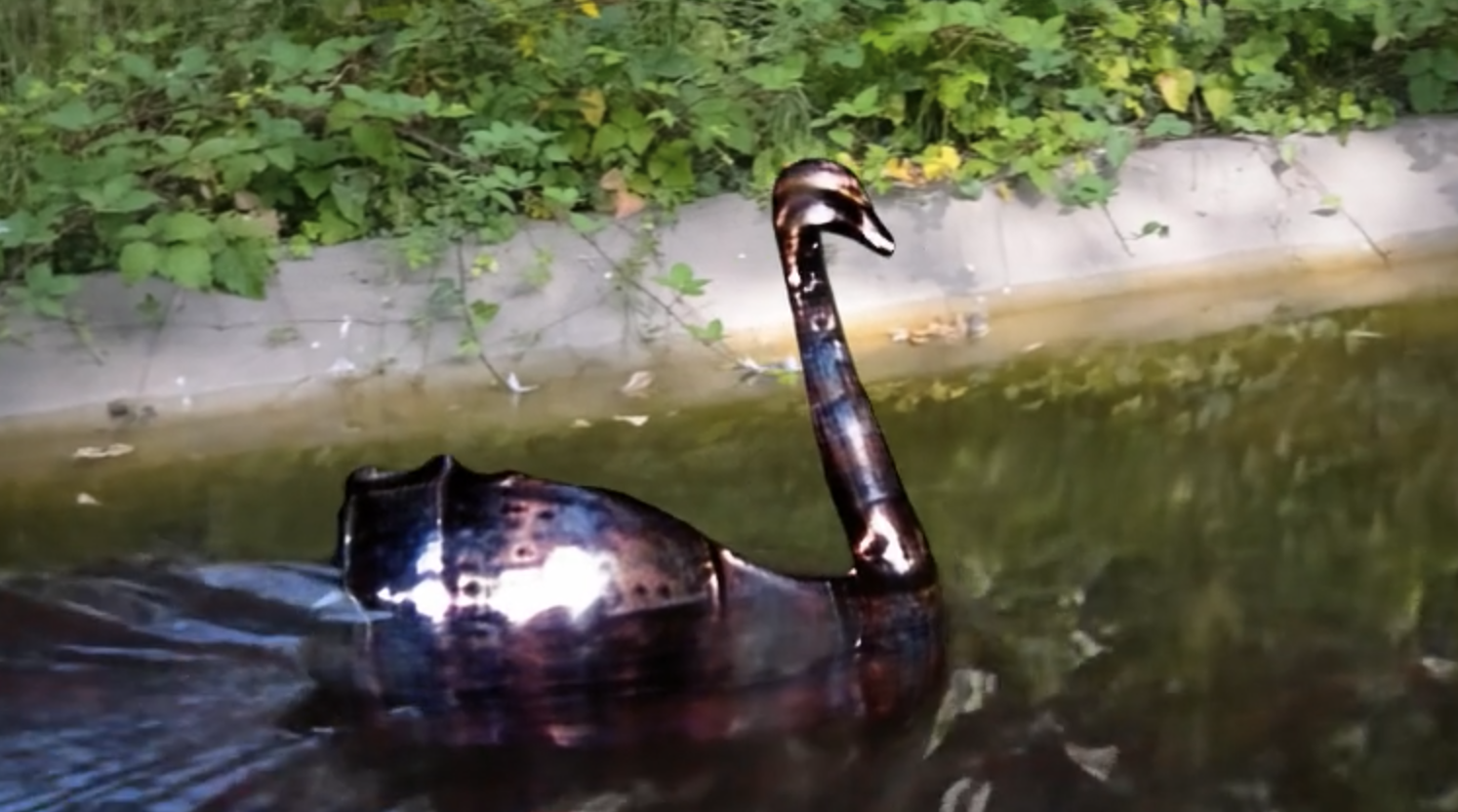} &
         \includegraphics[width=0.32\linewidth]{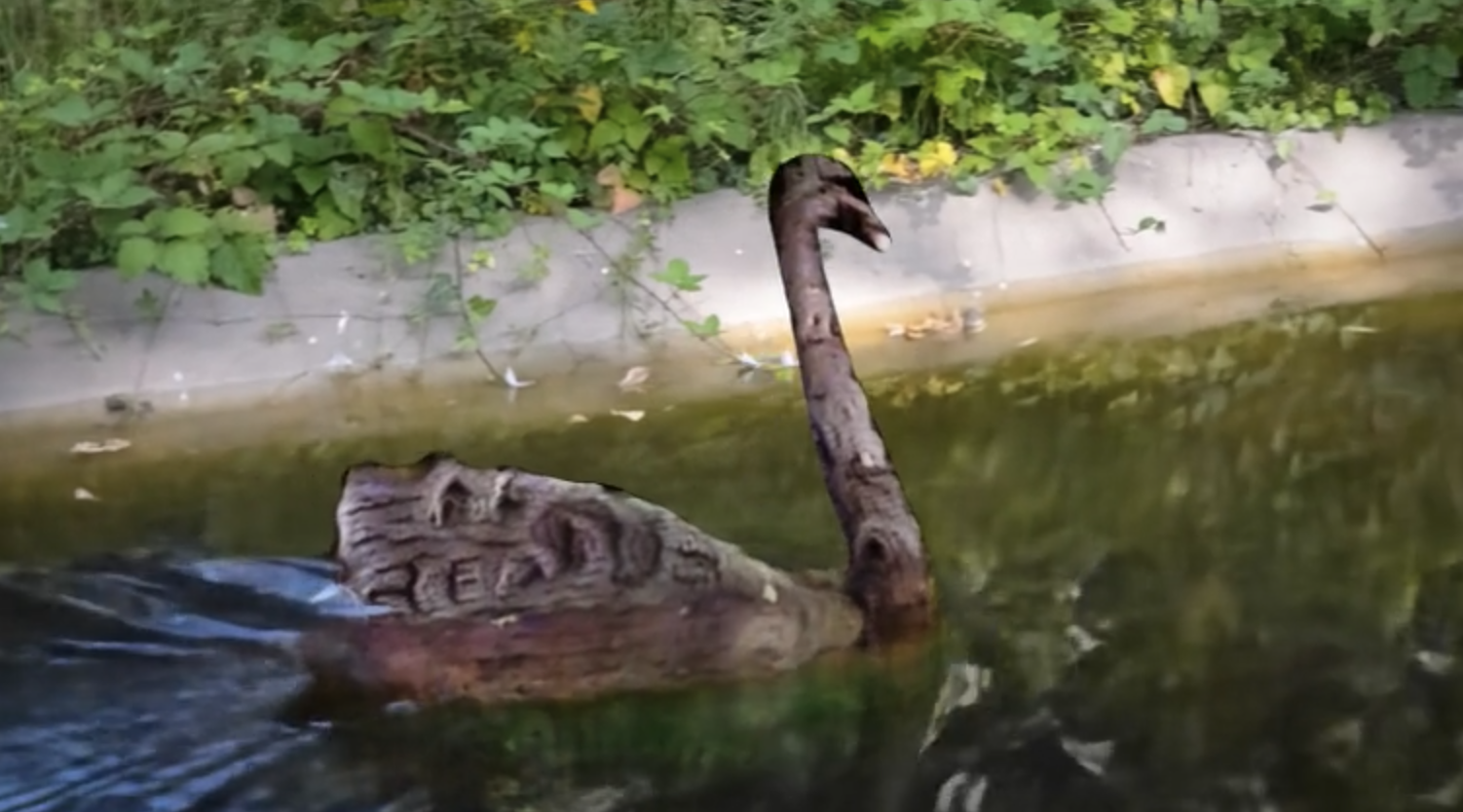} \\
         
         \includegraphics[width=0.32\linewidth]{images/example_images/org_swan2.png} &
         \includegraphics[width=0.32\linewidth]{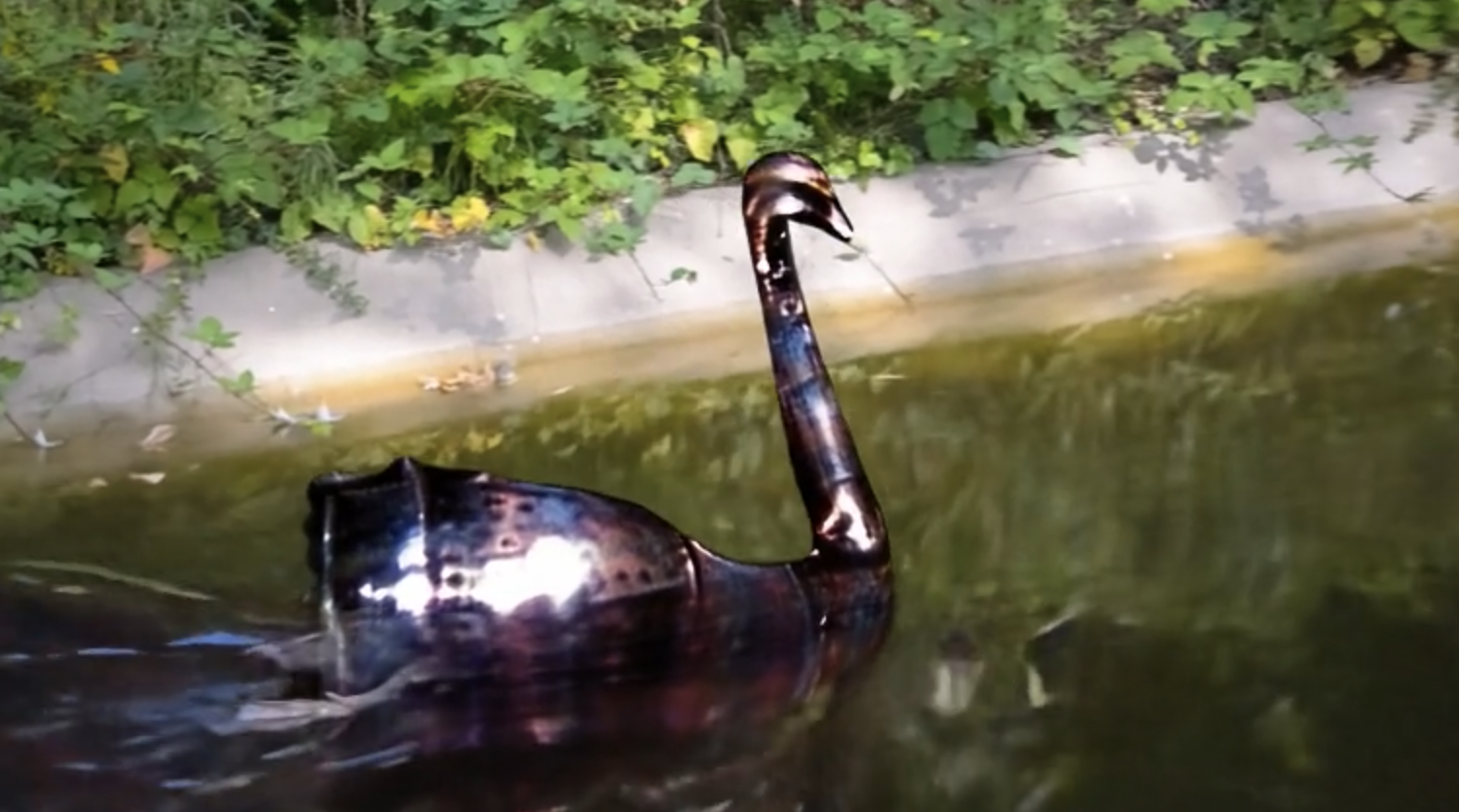} &
         \includegraphics[width=0.32\linewidth]{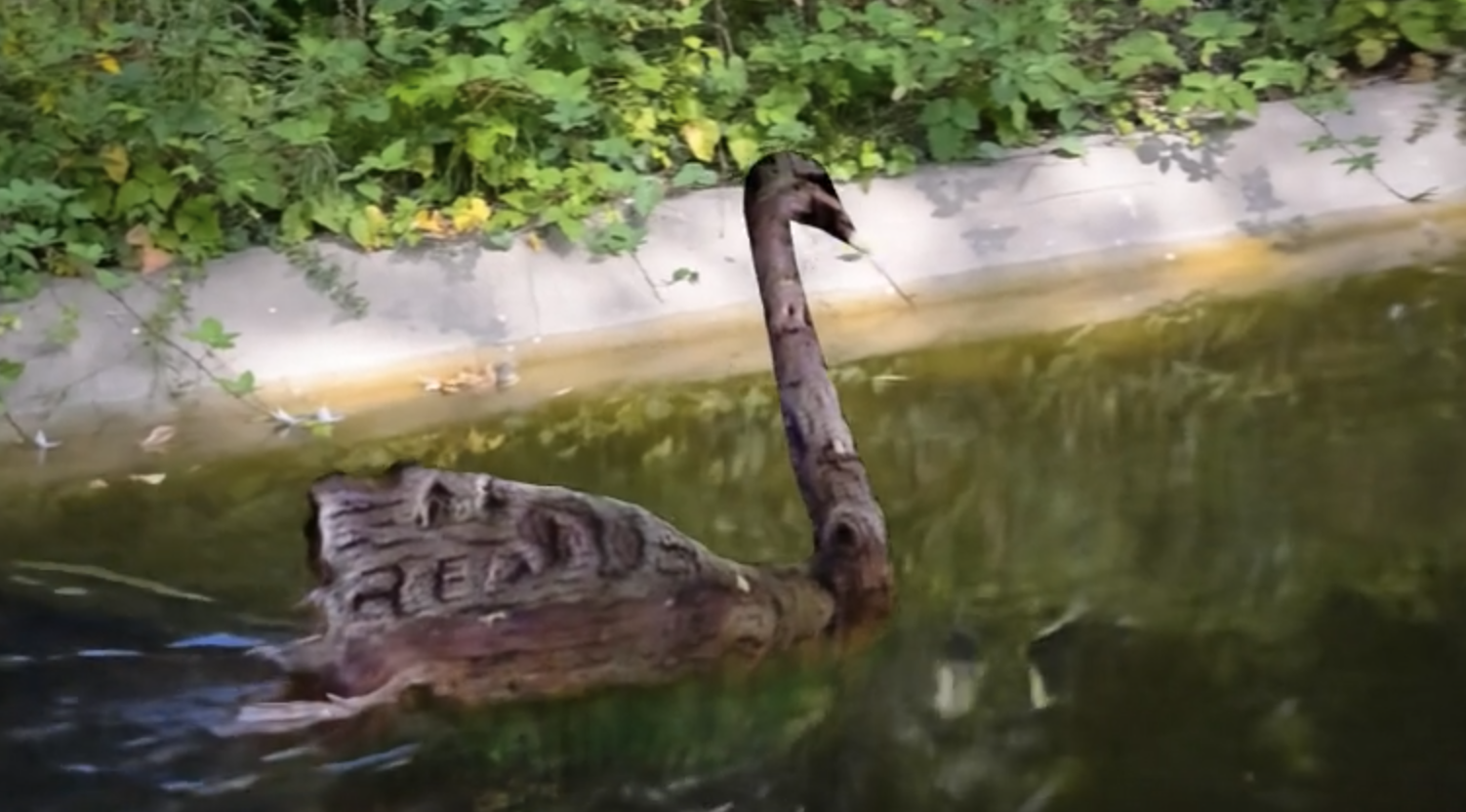} \\
         ``Swan'' &``Shiny metal swan'' &``Swan with wood bark skin'' \\
         \includegraphics[width=0.32\linewidth]{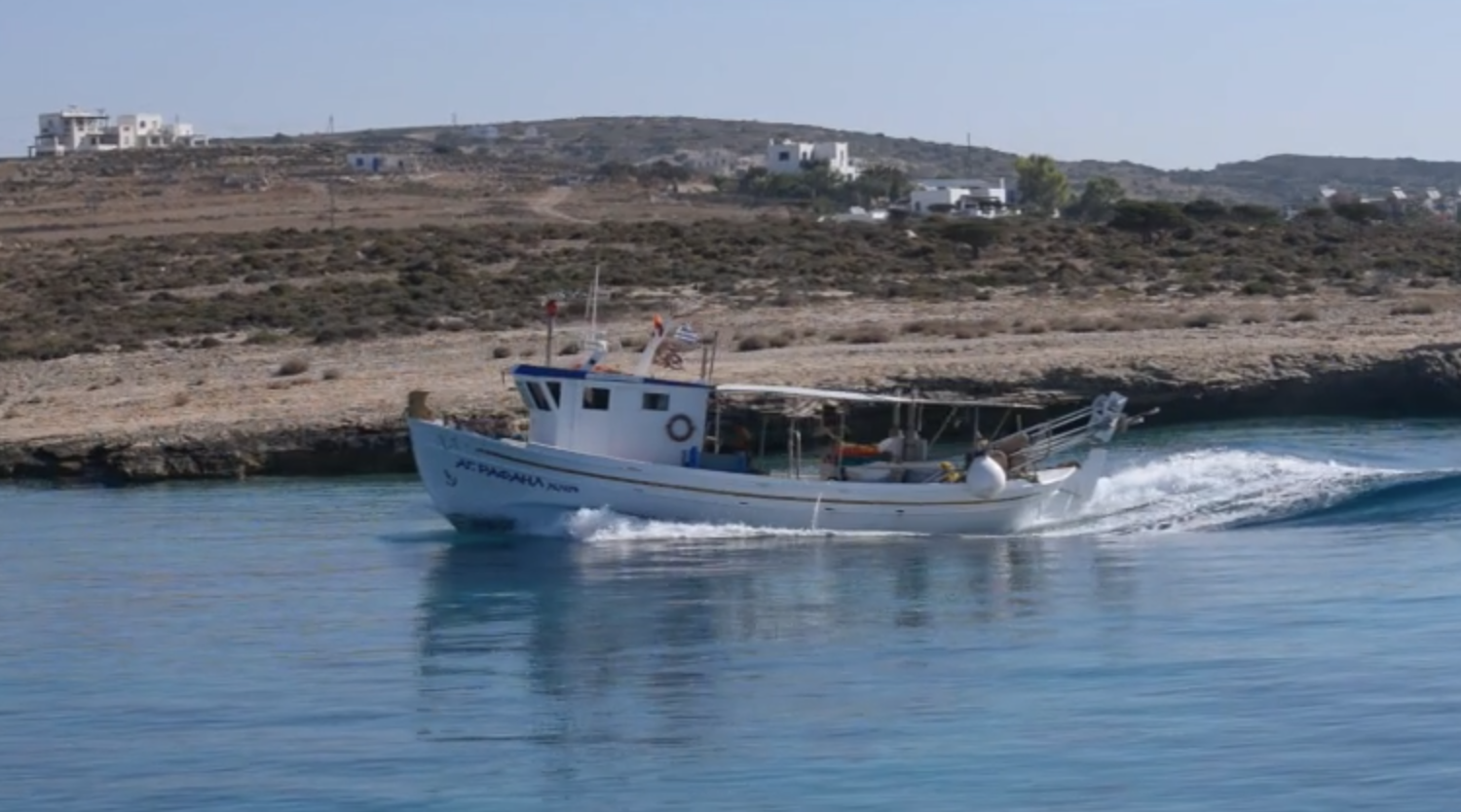} &
         \includegraphics[width=0.32\linewidth]{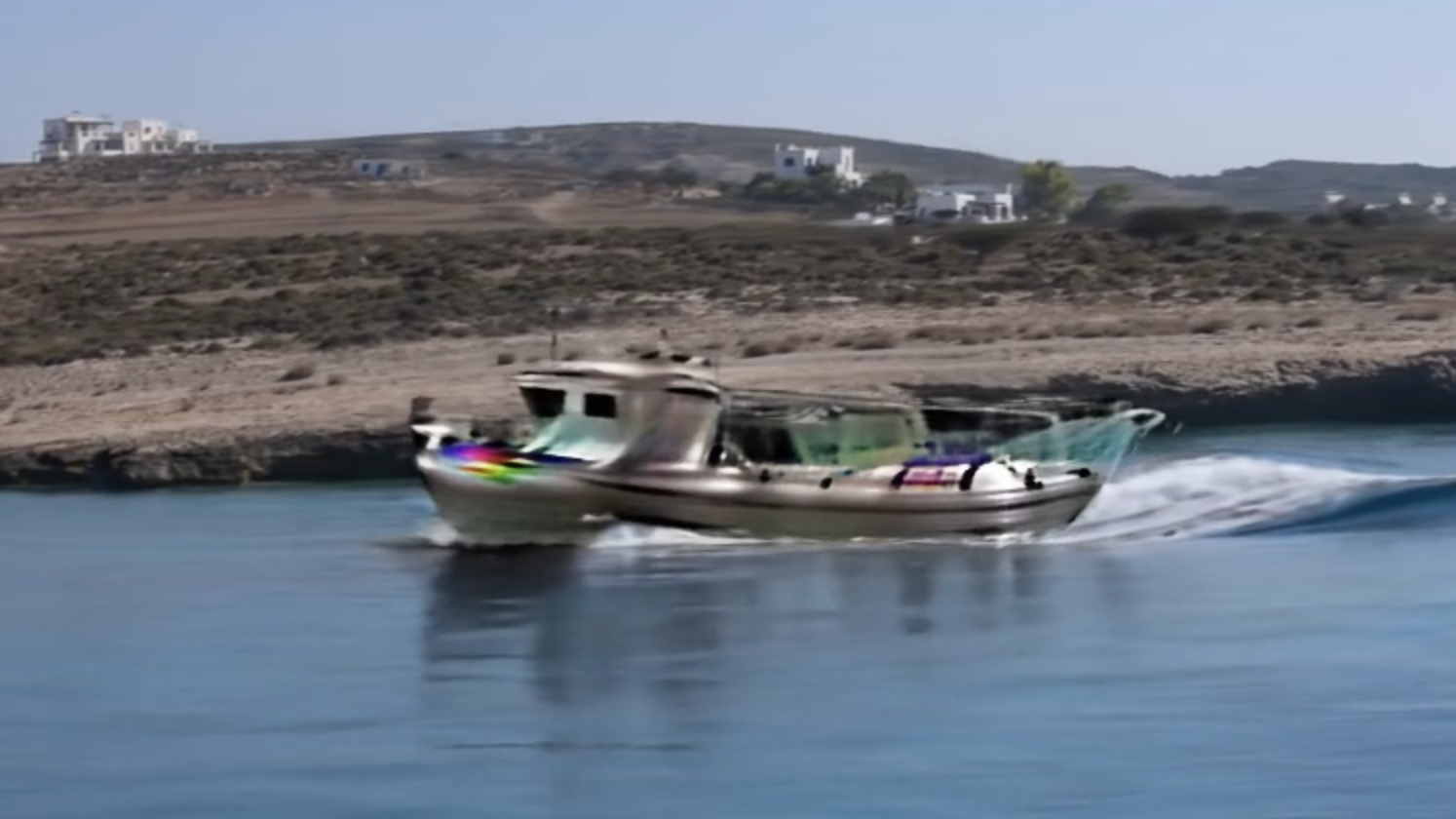} &
         \includegraphics[width=0.32\linewidth]{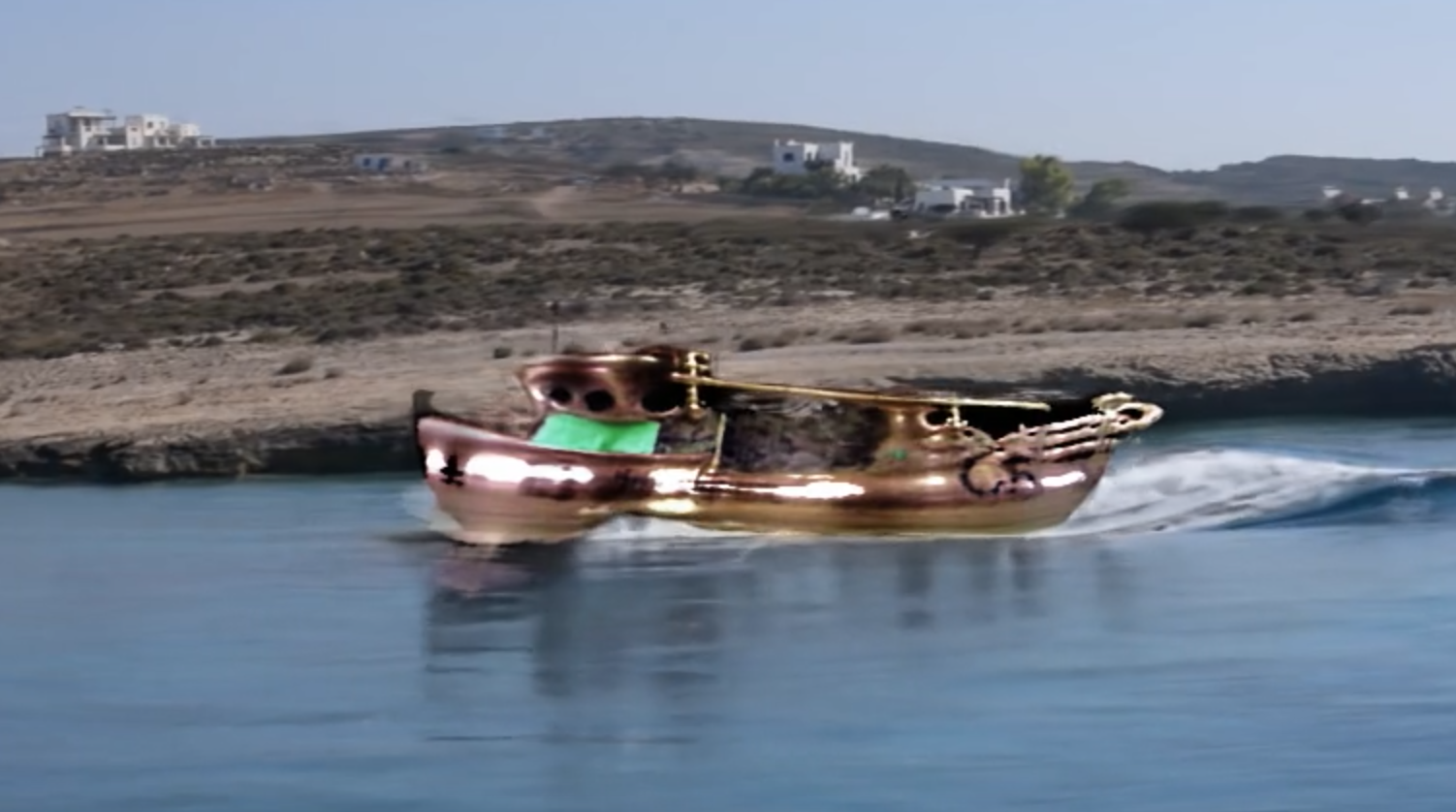} \\
         
         \includegraphics[width=0.32\linewidth]{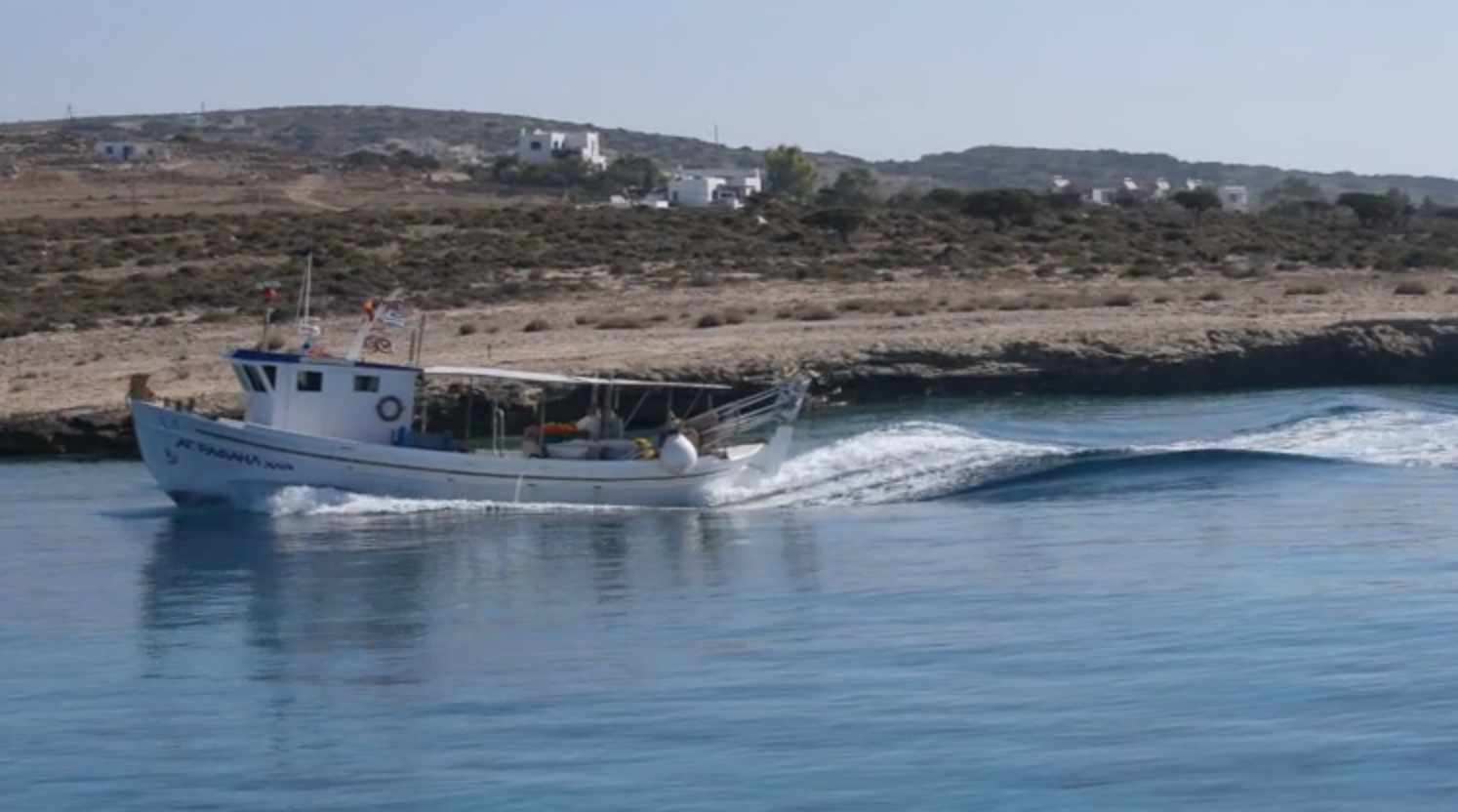} &
         \includegraphics[width=0.32\linewidth]{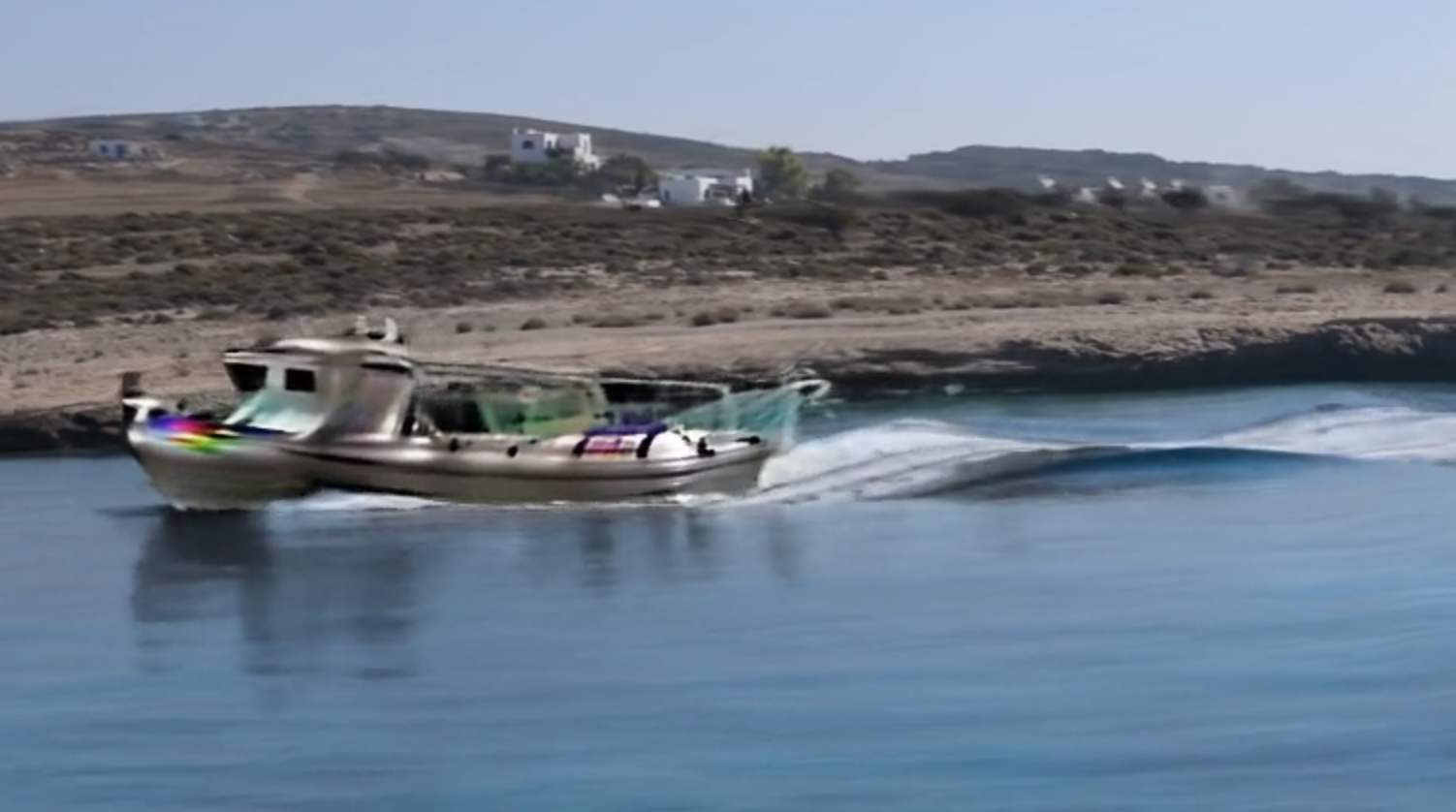} &
         \includegraphics[width=0.32\linewidth]{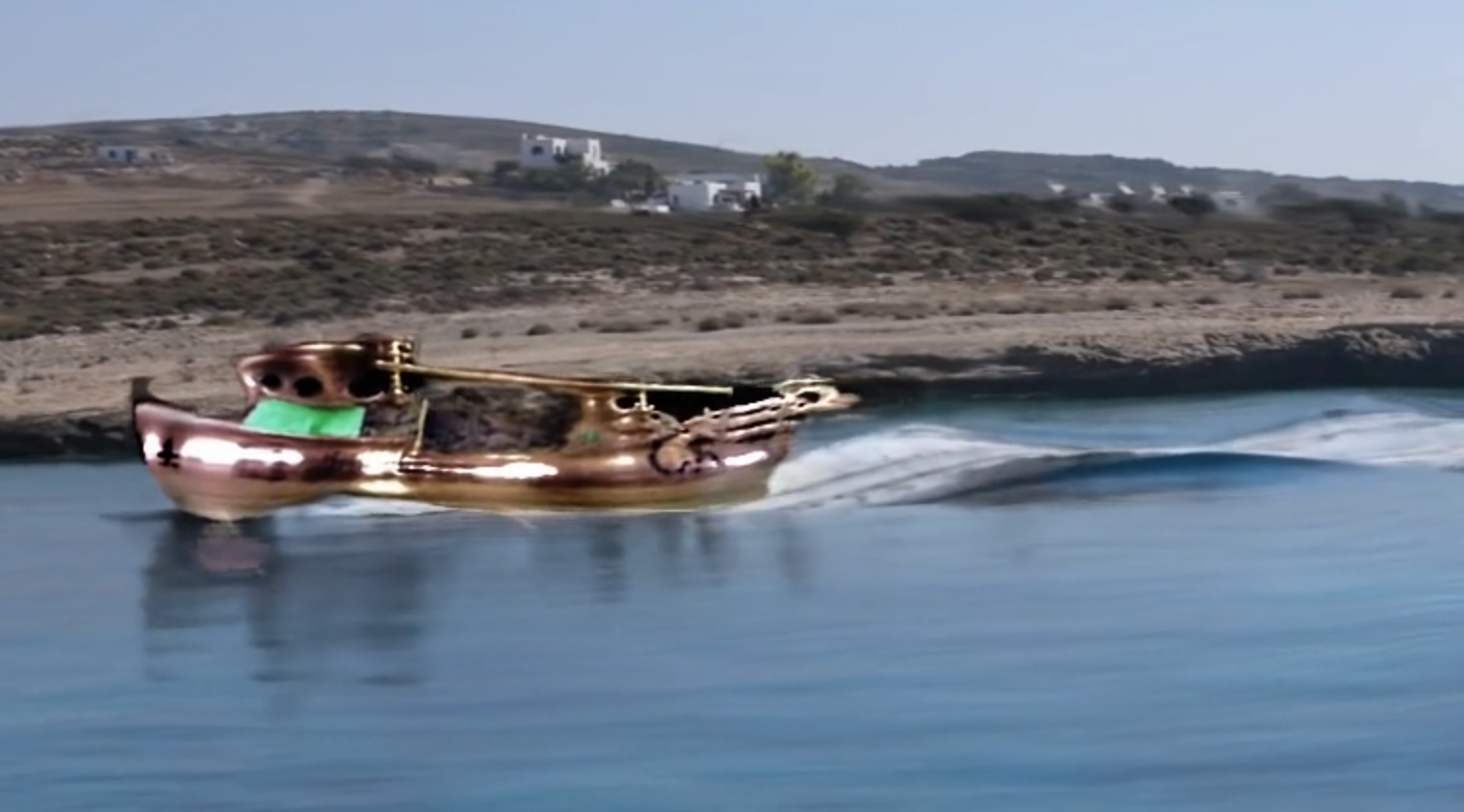} \\
         ``Boat'' &``Shiny aluminum fishing boat'' &``Boat made of copper'' \\
         \includegraphics[width=0.32\linewidth]{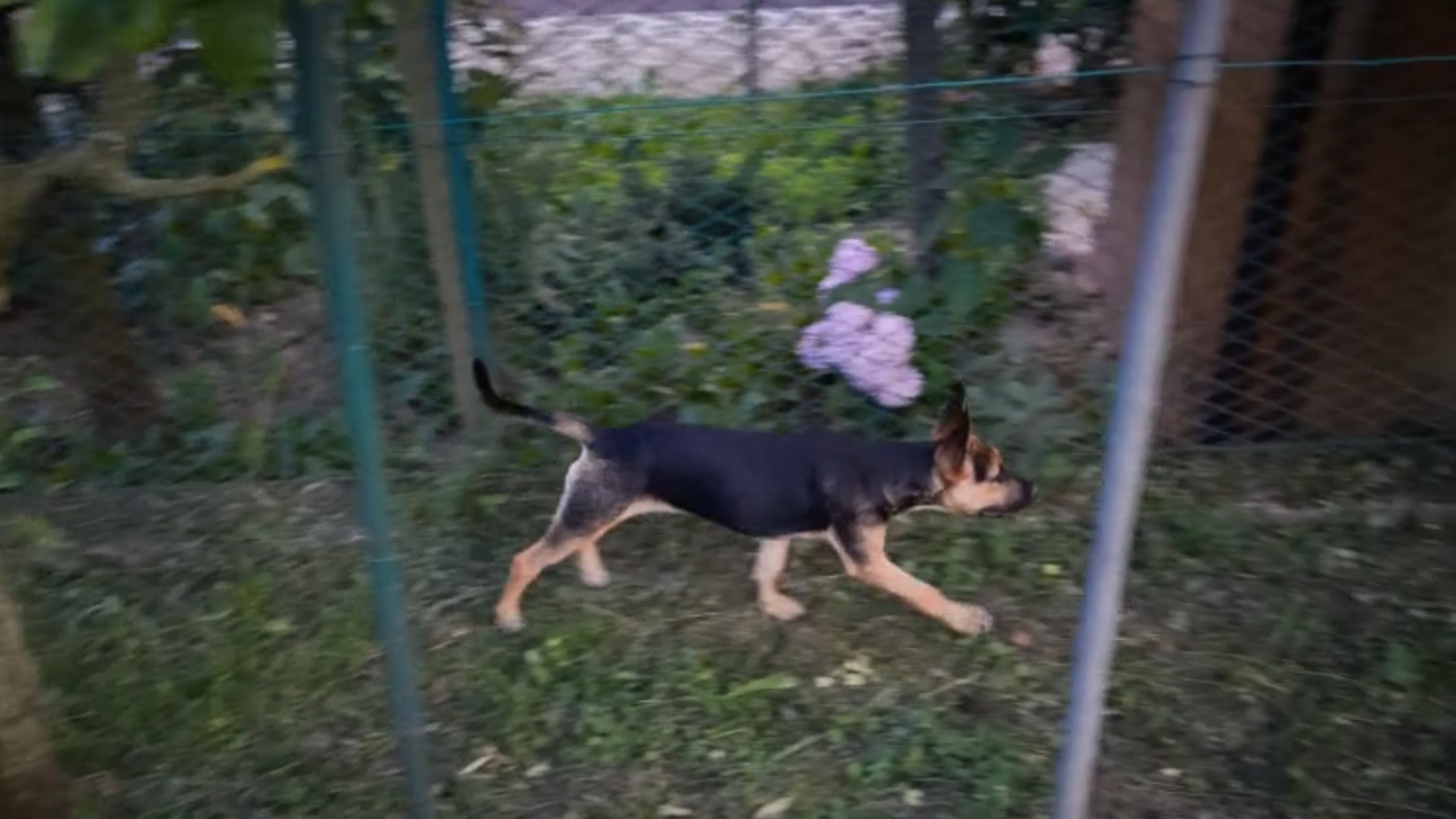} &
         \includegraphics[width=0.32\linewidth]{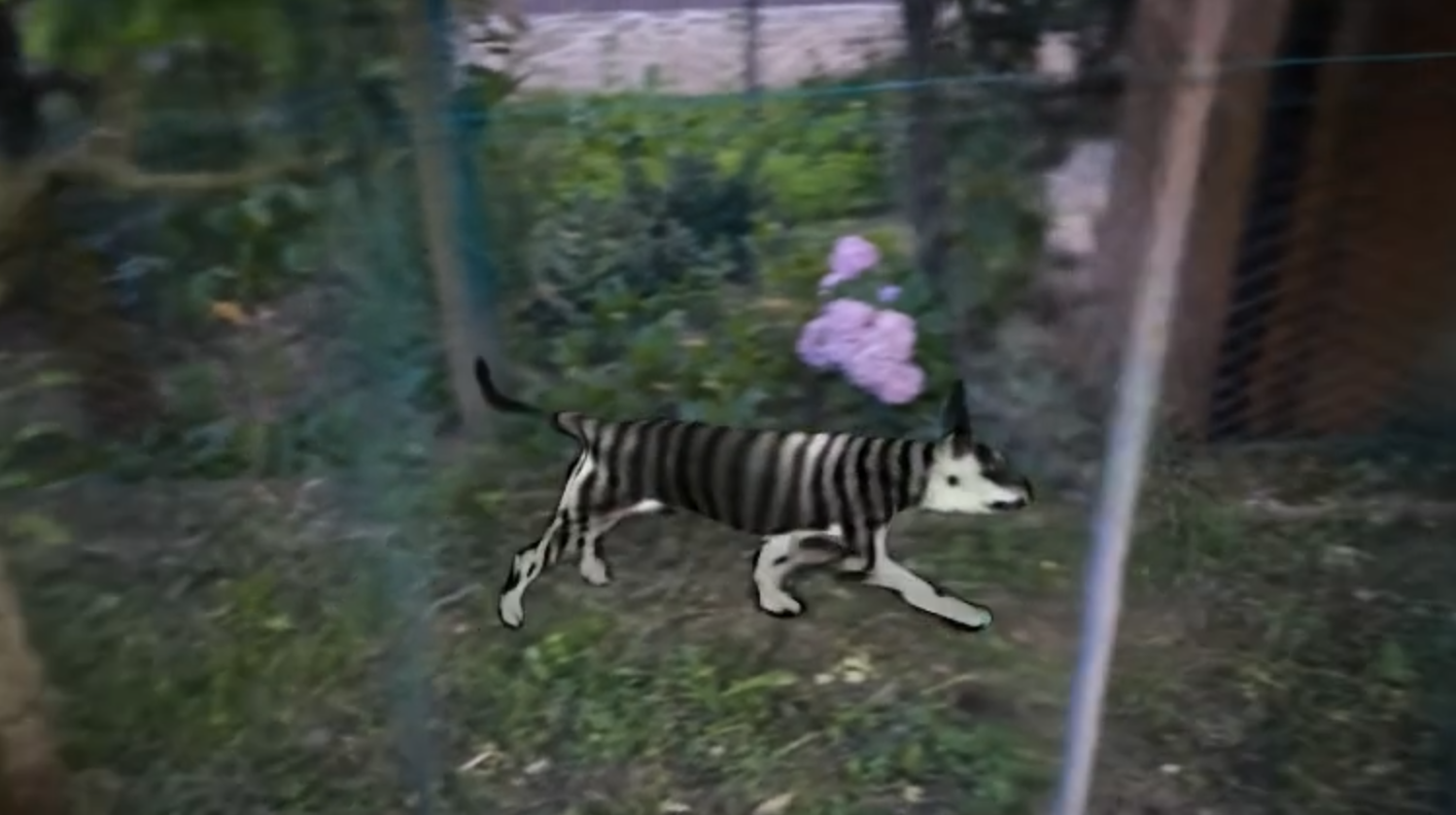} &
         \includegraphics[width=0.32\linewidth]{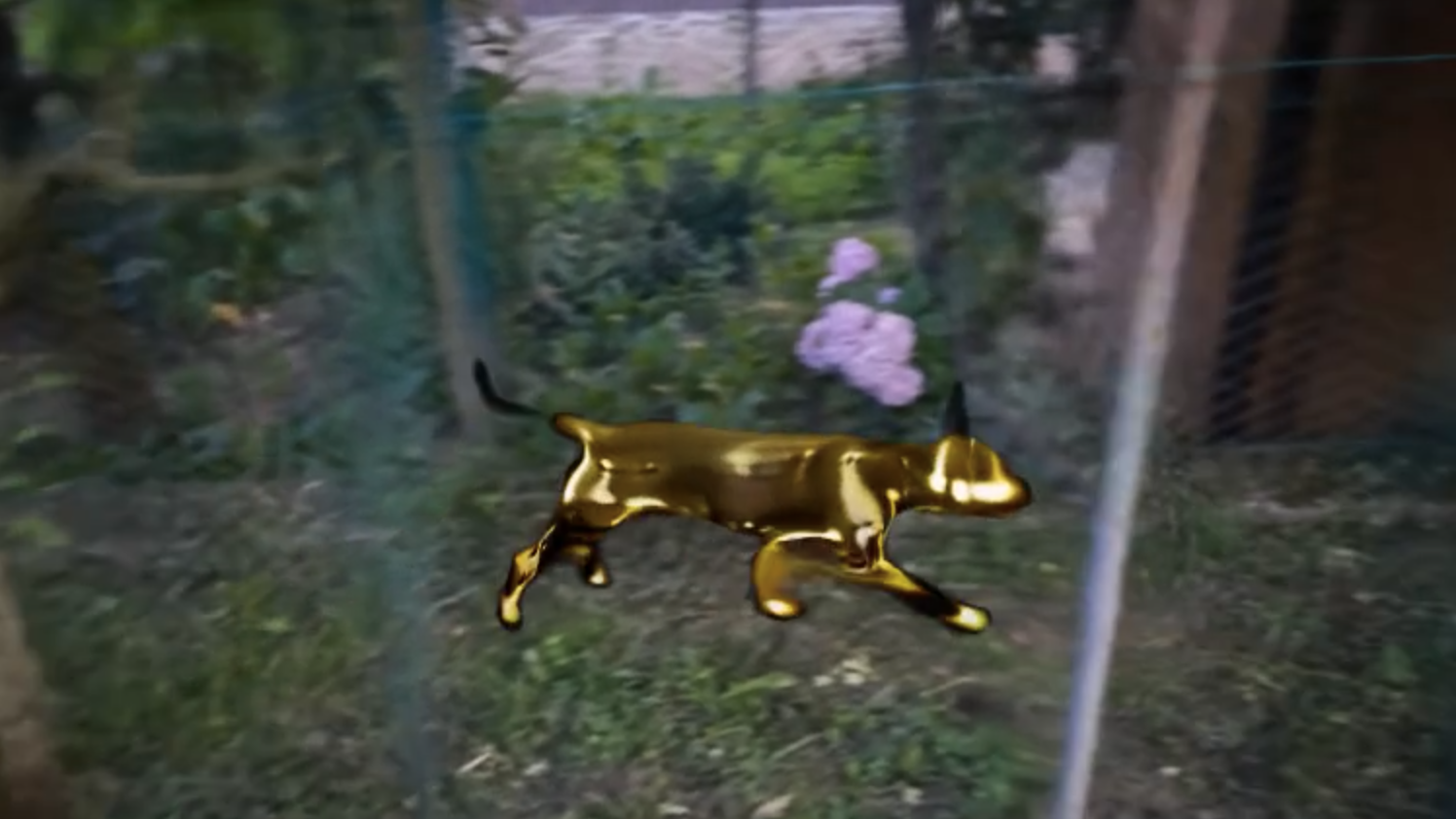} \\
         
         \includegraphics[width=0.32\linewidth]{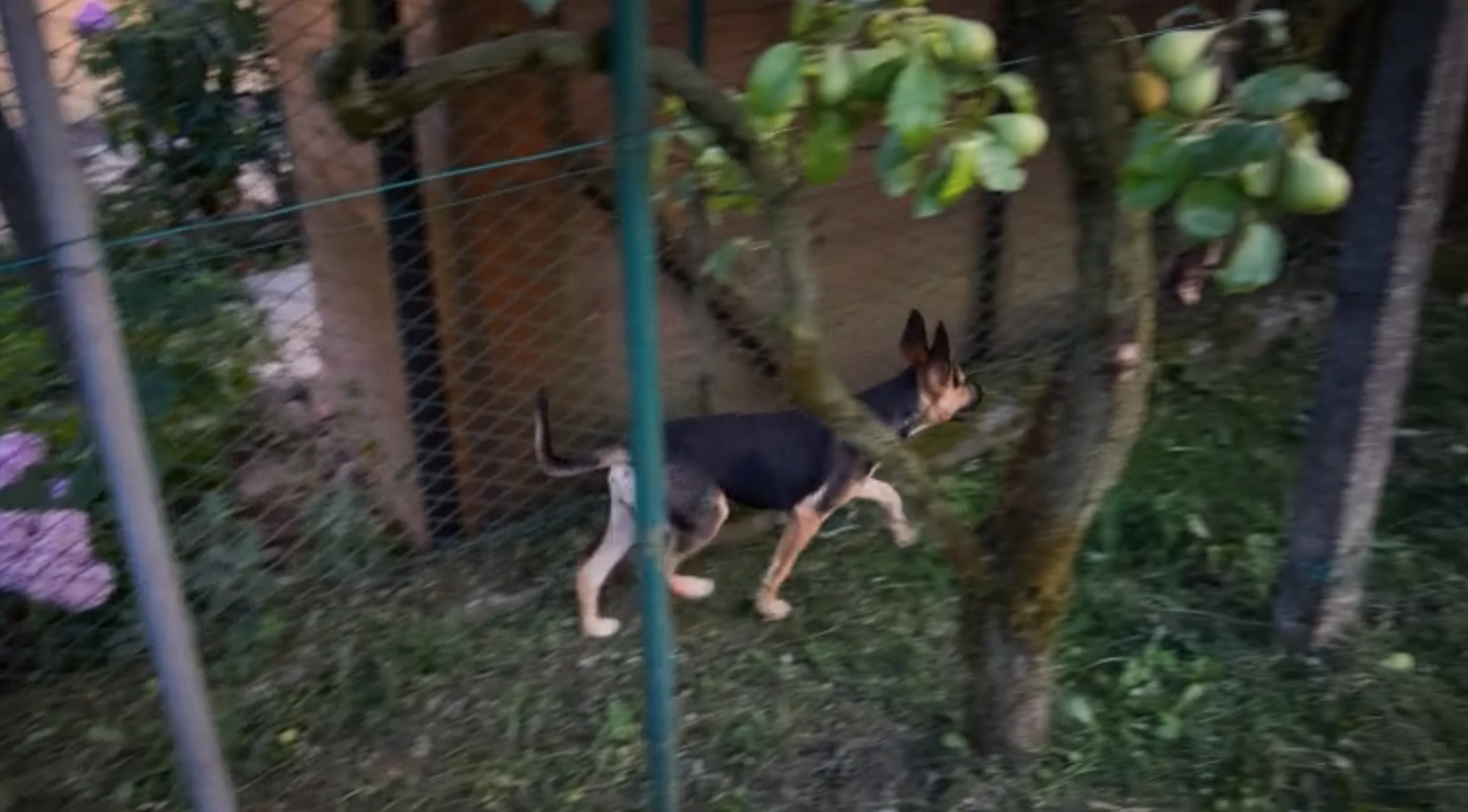} &
         \includegraphics[width=0.32\linewidth]{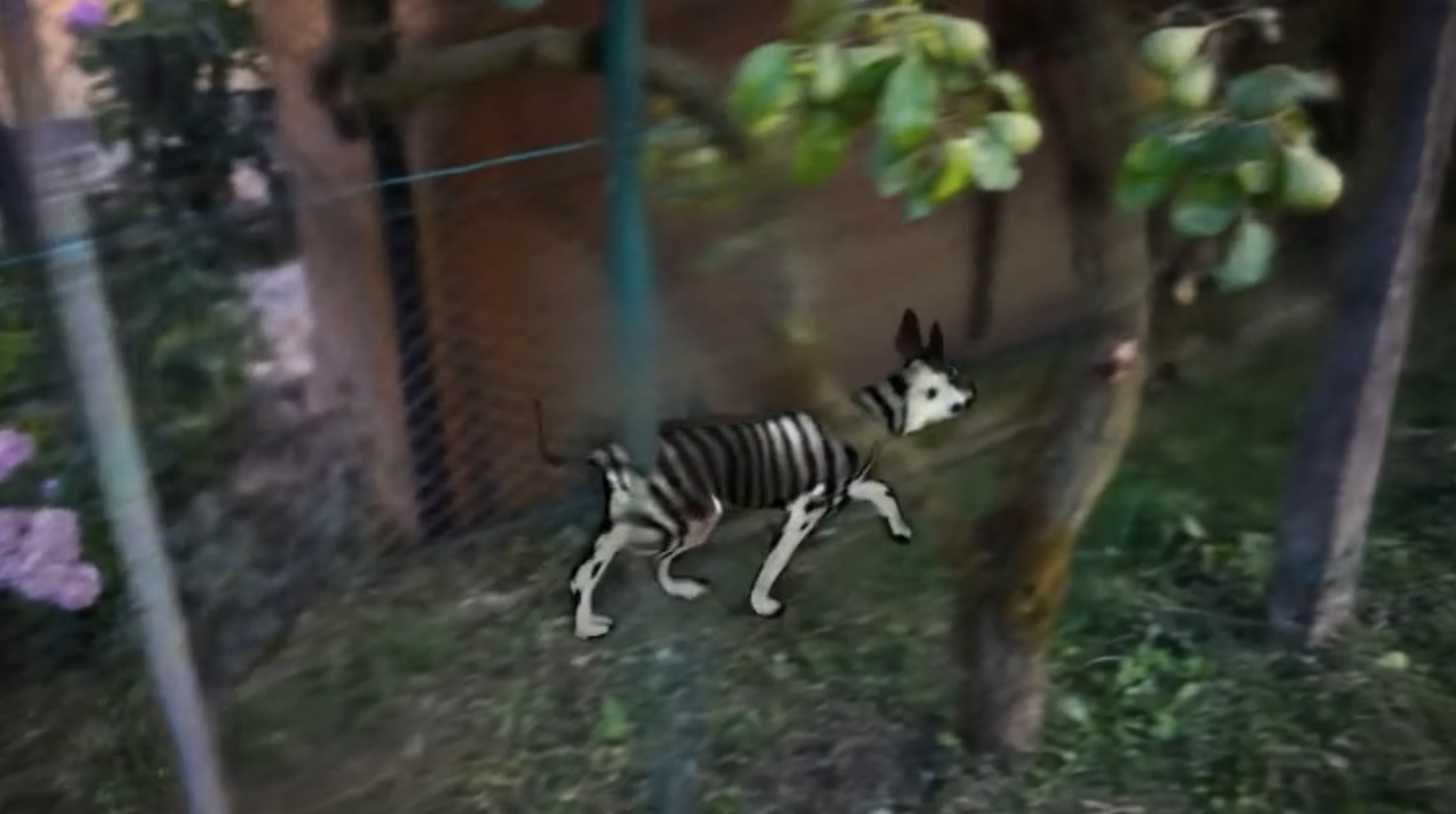} &
         \includegraphics[width=0.32\linewidth]{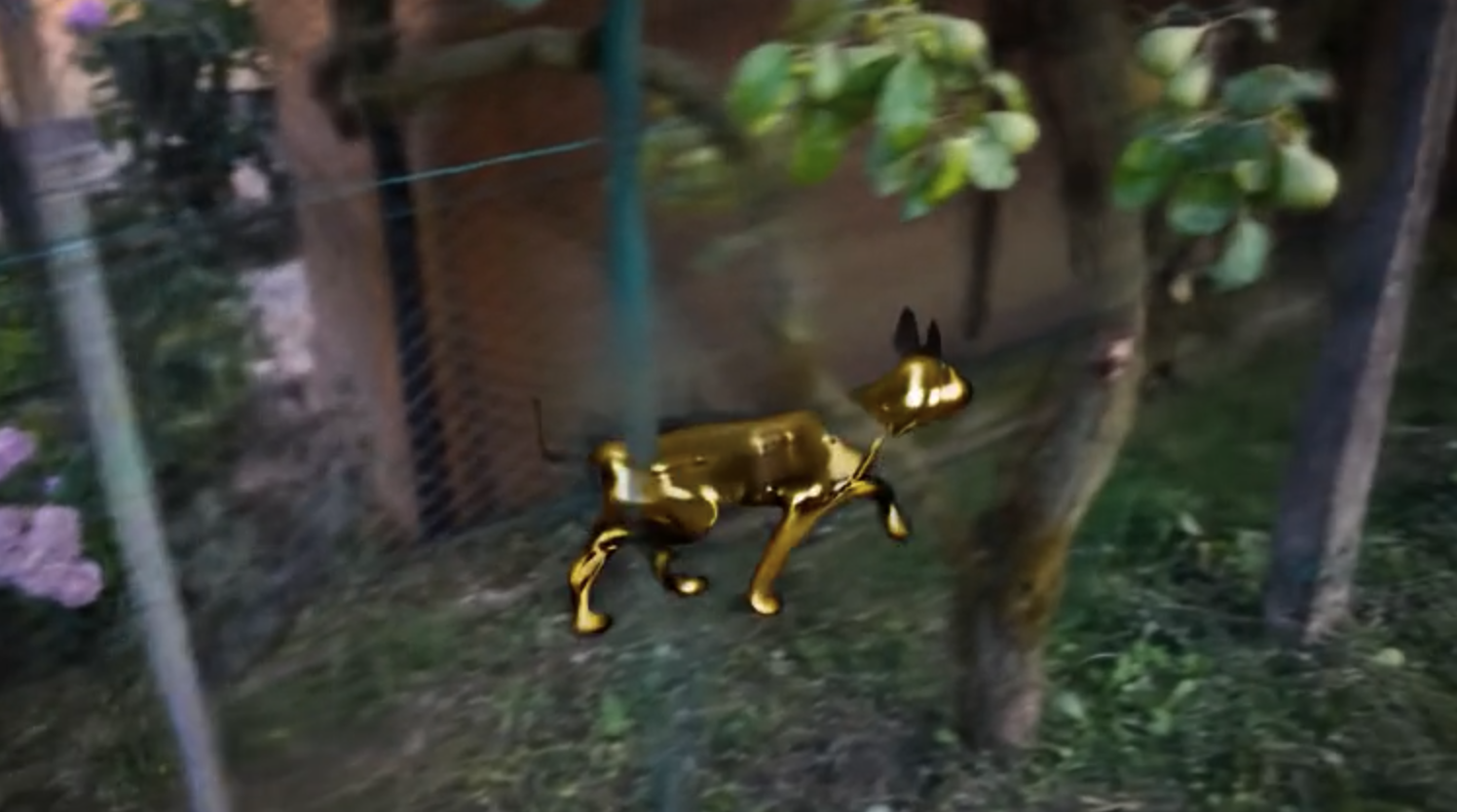} \\
         ``Dog'' &``Dog with zebra fur'' &``Golden dog'' 
         
\end{tabular}
     \caption{\textbf{Example results}. Two representative frames from each edited video together with the global target text.} 
    \label{fig:varity_examples}
\end{figure*}

\begin{table}[]
\centering
\begin{tabular}{lll}
\toprule
                             & Q1 (Realism) & Q2 (Matching Text) \\ \midrule
Blended Diffusion~\cite{blended_diffusion} &   $2.22$ ($\pm 1.08$)           &  $2.10$ ($\pm 1.00$)                    \\
Ours                   &   $\mathbf{3.47}$ ($\pm 1.10$)              &   $\mathbf{3.94}$ ($\pm 0.99$)                \\ \bottomrule
\end{tabular}
\caption{Mean opinion scores (1-5) and standard deviation for Q1 and Q2. \label{Tab:study_table}}
\end{table}

\subsection{Varied stylizations}\label{section:general_results}
In Fig.~\ref{fig:varity_examples} we illustrate our video stylization method applied to three videos and three texts. All local target texts used for these examples are similar to the global target text, but without containing the information about the underlying content, e.g., for the swan, the local target text is ``metal skin.'' The results show that we can apply temporally consistent stylizations that adhere to the target text specification. The swan example shows fine-grained details of the target texts and also preserves the underlying content of the swan. In the boat example, the stylization captures the details of the target text. The boat has a texture similar to shiny aluminum and also has something that looks like a fishing net at the end of the boat. The dog example shows that our method can apply a realistic and consistent stylization to a video containing occlusions.

We quantify the effectiveness of our method by comparing it to an image baseline applied to each frame in a video input. As a baseline, we use a pretrained Blended-diffusion (BF)~\cite{blended_diffusion} model with standard configurations. The model takes as input an image, a ROI mask, and a target text. BF performs local (region-based) edits based on a target text description and the ROI mask.
We conduct a user study to evaluate the perceived quality of the stylized outputs generated by both the BF model and our method, and the degree to which the outputs adhere to the global target text.
Our user study comprises $50$ users and $20$ stylized videos, each with a different target text.  For each video and target text combination, the users are asked to assign a score ($1$-$5$) to two factors: (Q1) ``How realistic is the video?,'' (Q2) ``does the \{\textit{object}\} in the video adhere to the text \{\textit{content}\}. For Q1 we make it clear to the user that ``realistic'' refers to the quality of the video content. The results are shown in Table~\ref{Tab:study_table}.

\begin{figure}[H]
\centering
\begin{tabular}{ccccc}
        
         \includegraphics[width=0.191\linewidth] {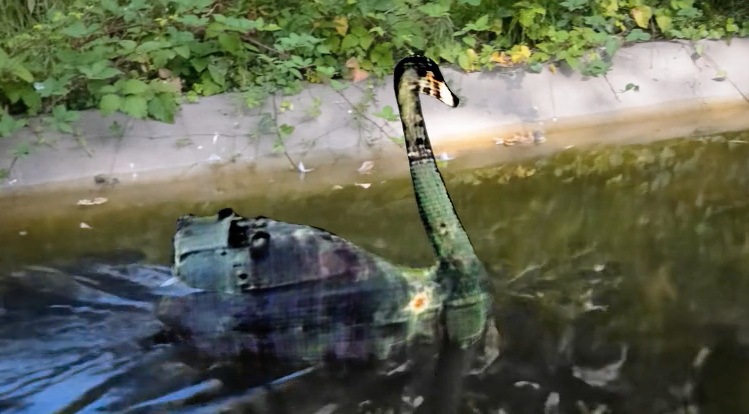} &
         \includegraphics[width=0.191\linewidth]{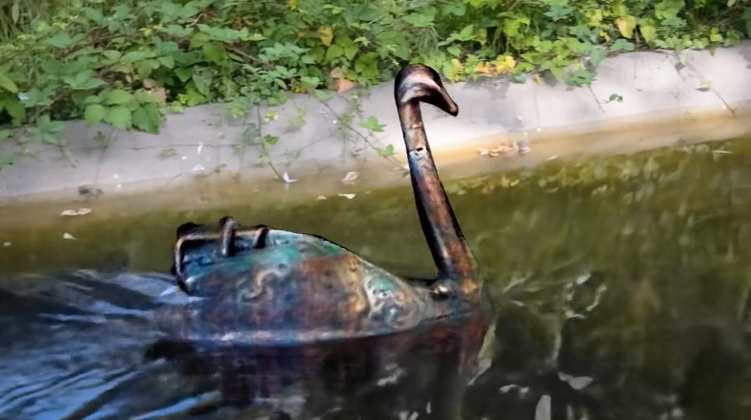} &
         \includegraphics[width=0.191\linewidth]{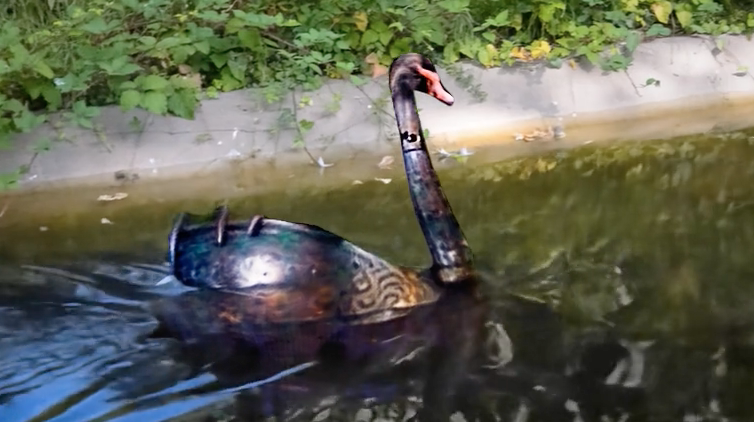} &
         \includegraphics[width=0.191\linewidth]{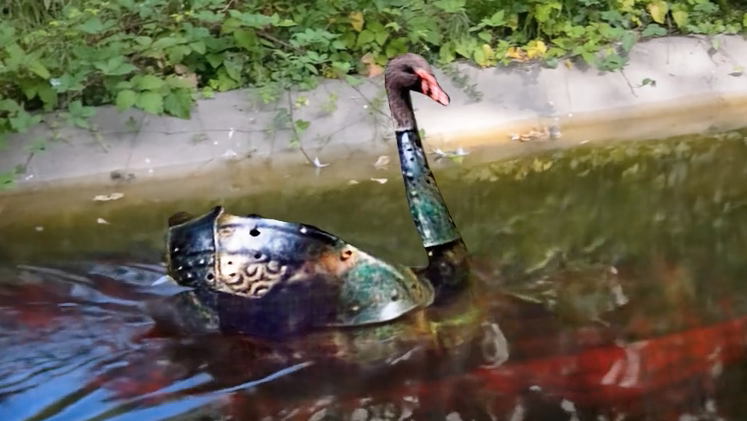} &
         \includegraphics[width=0.191\linewidth]{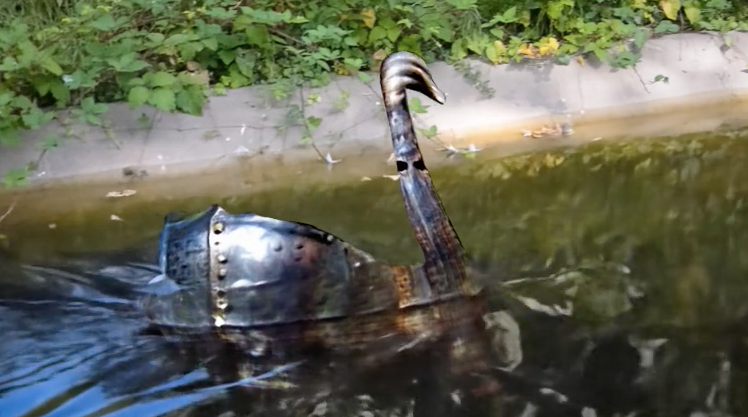} 
         \\
         (a) & (b) &(c) &(d) &(e) 
\end{tabular}

\begin{tabular}{ccccc}
        
         \includegraphics[width=0.191\linewidth]{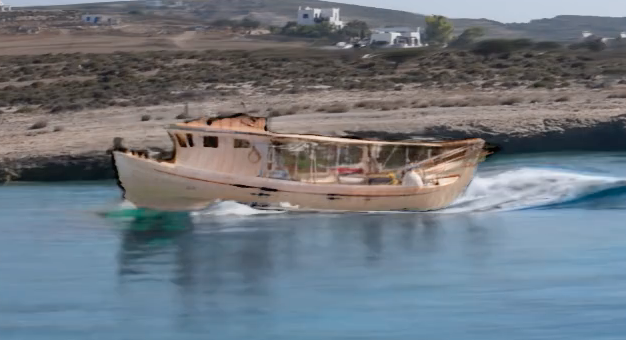} &
         \includegraphics[width=0.191\linewidth]{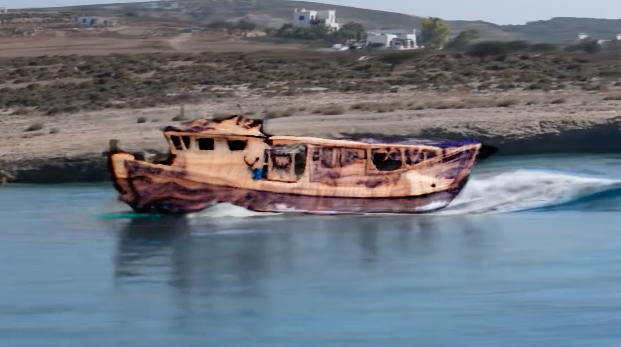} &
         \includegraphics[width=0.191\linewidth]{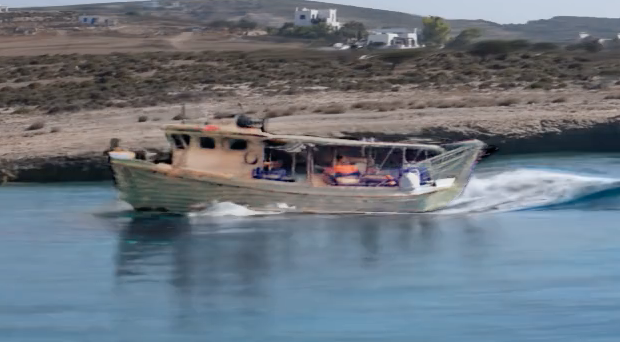} &
         \includegraphics[width=0.191\linewidth]{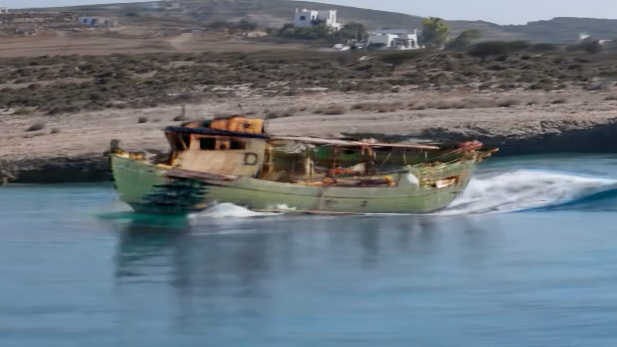} &
        \includegraphics[width=0.191\linewidth] {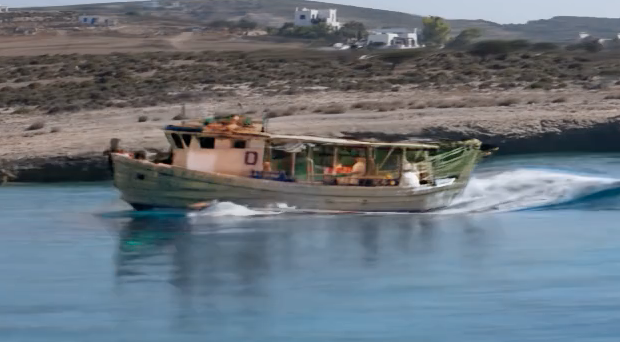}\\ 
        (a) & (b) &(c) &(d) &(e) 
\end{tabular}
     \caption{\textbf{Target text specificity.} Each example shows a representative frame from the video. The experiment shows the specificity of a target affects the stylization.
     Global target text prompts, \textit{row one:} (a) ``Armor,'' (b) ``Iron armor,'' (c) ``Medieval iron armor,'' (d) ``Suit of shiny medieval iron armor,'' (e) ``Full plate shiny medieval iron armor,'' \textit{row two:} (a) ``Boat made of wood,'' (b) ``Boat made of dark walnut wood (c) ``Fishing boat made of wood,'' (d) ``Old fishing boat made of wood,'' (e) ``Fishing boat made of wood planks.''} 
    \label{fig:specificity_exp}
\end{figure}

\subsection{Prompt specificity}\label{section:prompt_specificity}
We demonstrate that varying the specificity of the target text prompt affects the level of detail in the stylization. Our experiment is motivated by recent work on prompt engineering~\cite{LearningToPrompt} that shows how slight changes in the target text can have a big impact on the CLIP similarity between a text and an image. Fig.~\ref{fig:specificity_exp} shows an increasing level of detail for two videos and two target texts. The target text specificity not only influences the level of detail, it also makes it easier for CLIP to navigate in its embedding space. In the swan example in column (a), we have ``Swan with an armor,'' which is a more ambiguous target compared to the other swan examples with more detailed target texts and stylizations. We hypothesize that this is because several stylizations can satisfy the more simple target text, while a more specific target text narrows down the set of possible directions in CLIP's embedding space. The swan examples in columns (d) and (e) indicate that CLIP has some understanding of the different body parts of the swan. The ``full plate'' target text (d) covers the entire head of the swan while the ``suit'' of armor in column (e) has a clear cut around the head of the swan.
\begin{figure}
\vspace{-0.2cm}
\centering
\begin{tabular}{ccccc}
         \includegraphics[width=0.191\linewidth] {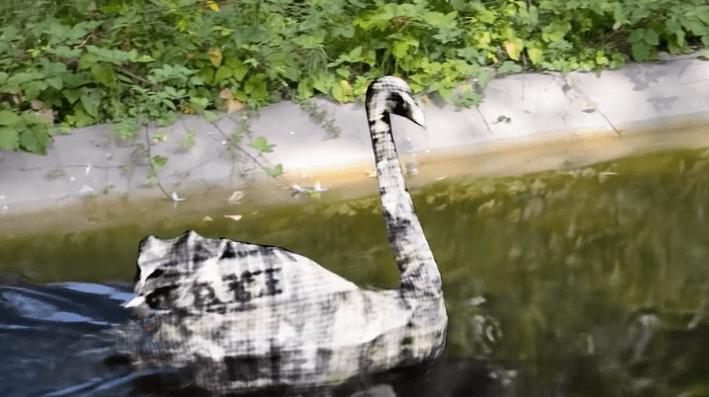} &
         \includegraphics[width=0.191\linewidth]{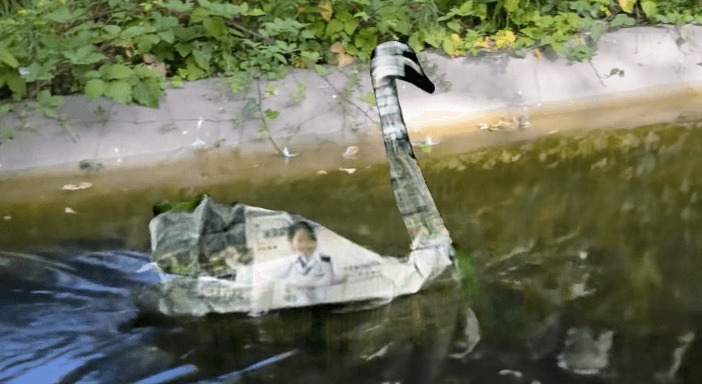} &
         \includegraphics[width=0.191\linewidth]{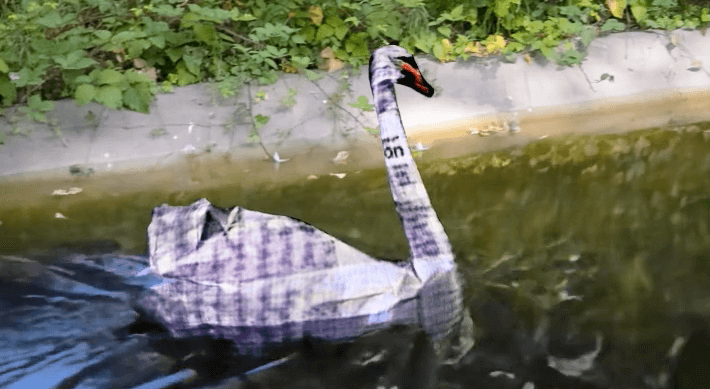} &
         \includegraphics[width=0.191\linewidth]{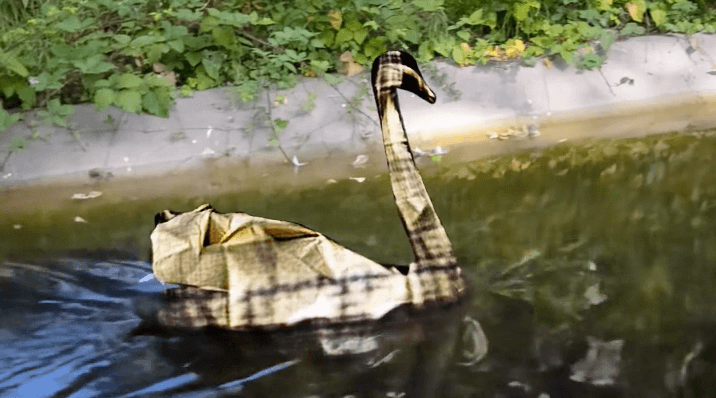} &
         \includegraphics[width=0.191\linewidth]{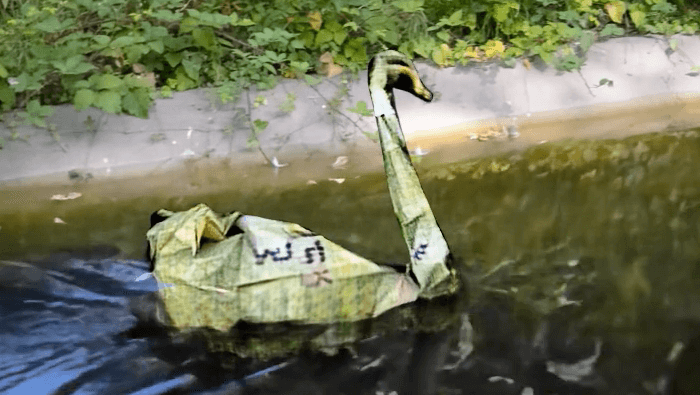} 
         \\
         (a) & (b) & (c) & (d) & (e)
\end{tabular}
\begin{tabular}{ccccc}
        
         \includegraphics[width=0.191\linewidth]{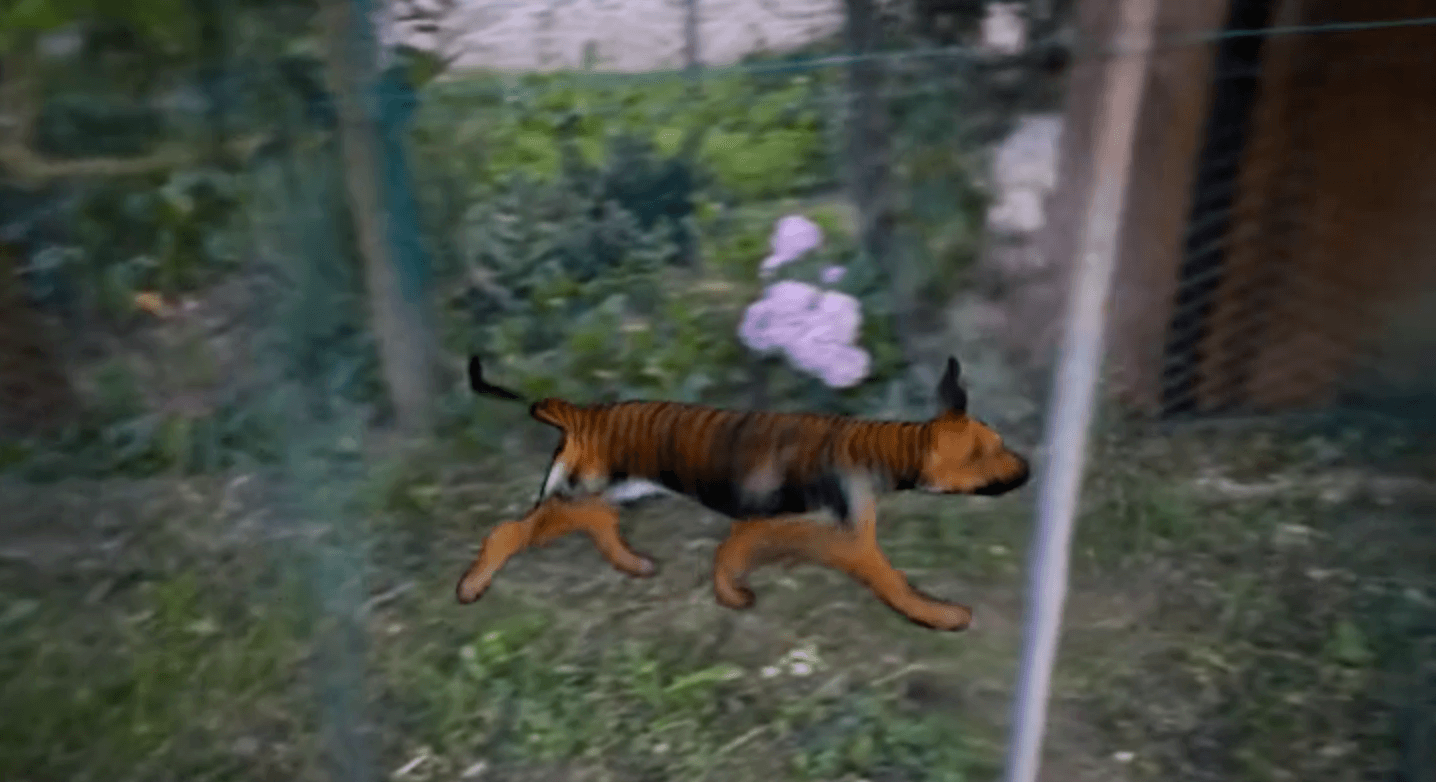} &
         \includegraphics[width=0.191\linewidth]{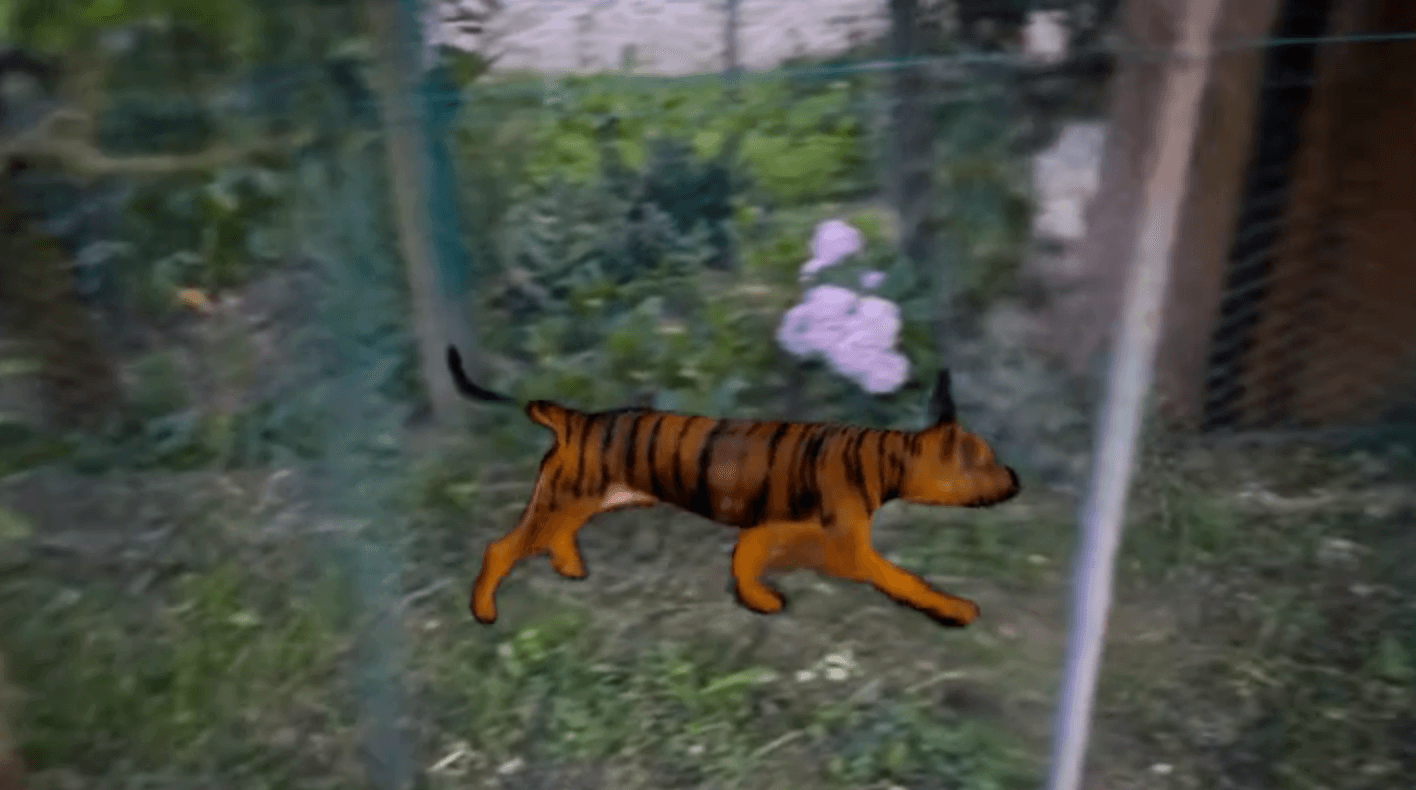} &
         \includegraphics[width=0.191\linewidth]{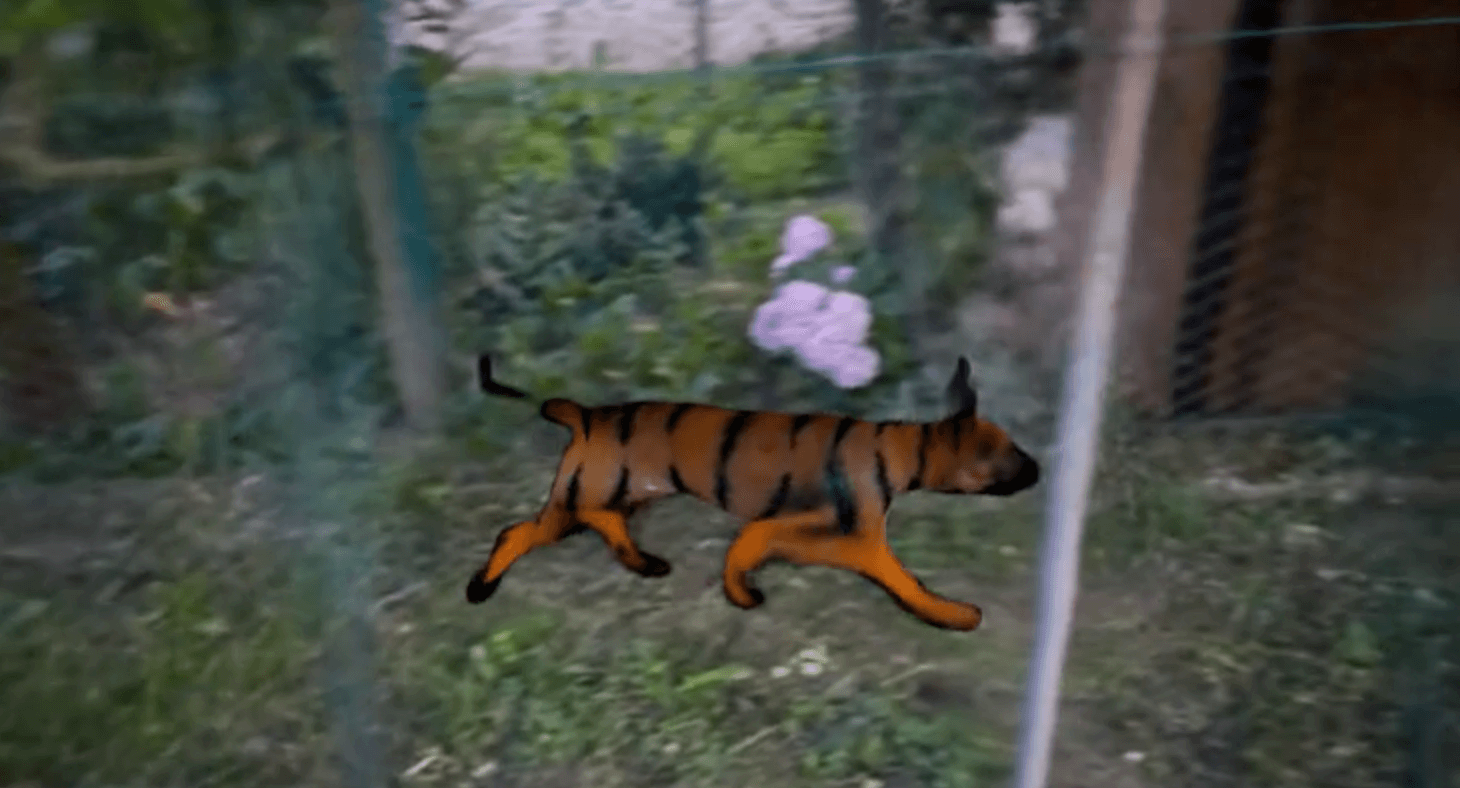} &
         \includegraphics[width=0.191\linewidth]{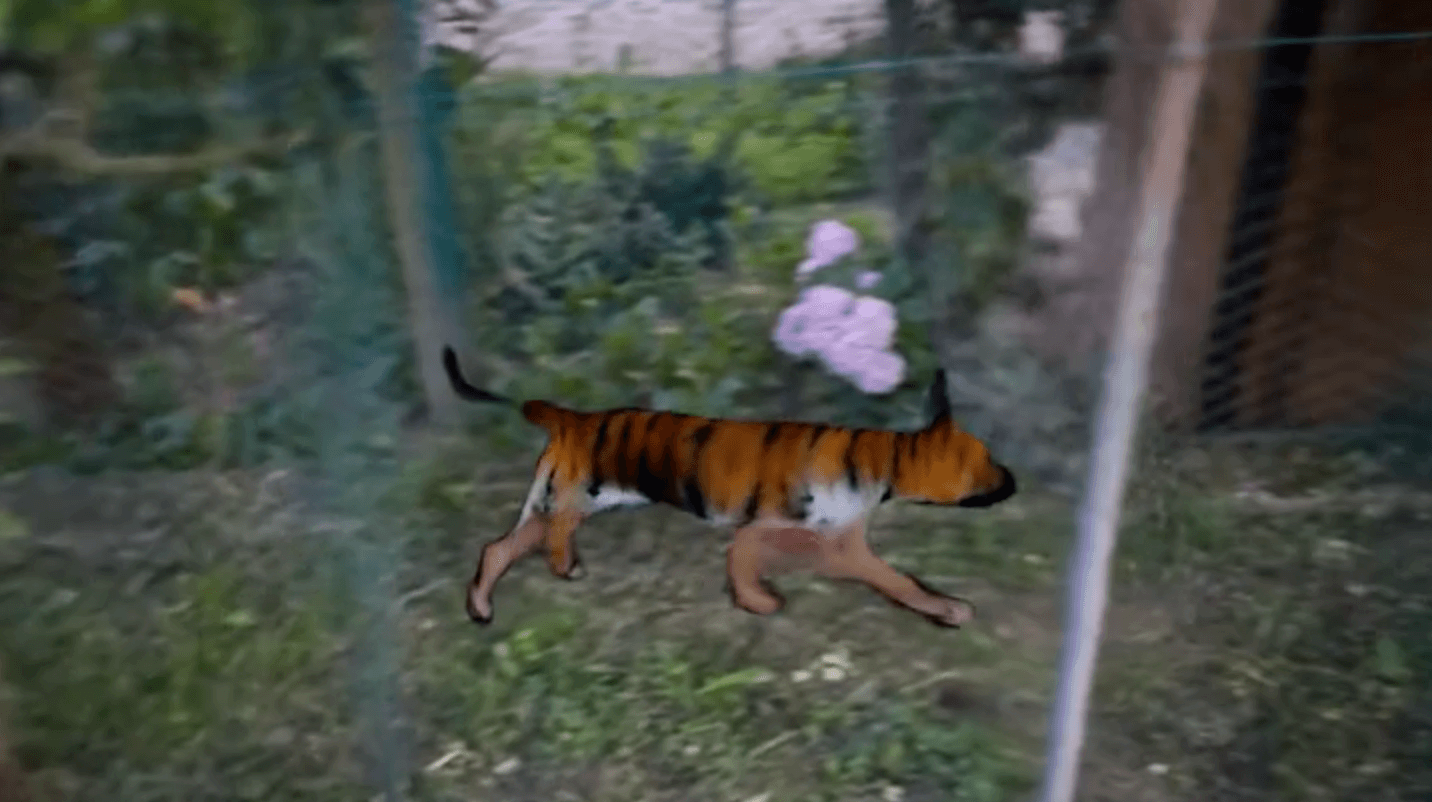} &
        \includegraphics[width=0.191\linewidth] {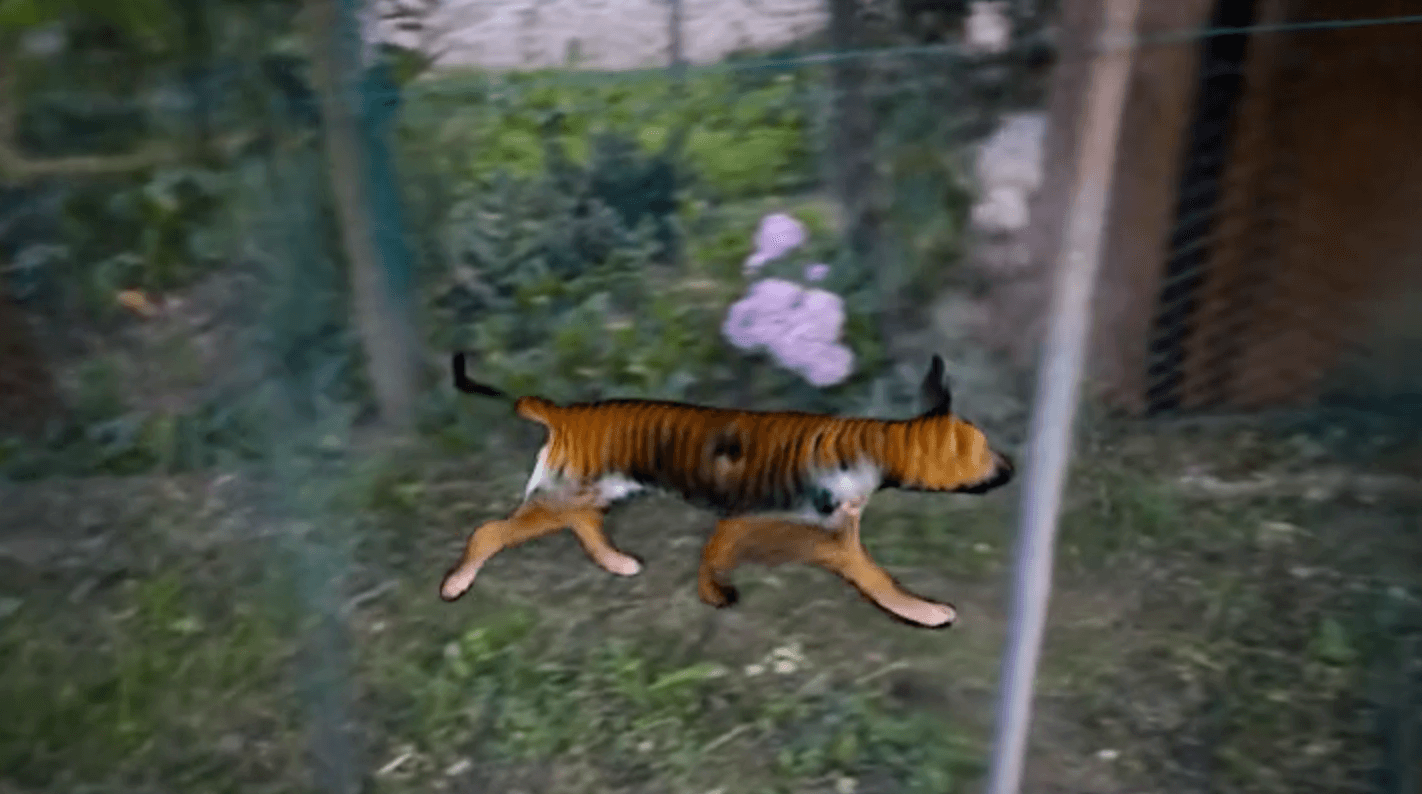} \\
        (a) & (b) &(c) &(d) &(e) 
\end{tabular}
     \caption{\textbf{Prefix augmentations - varying number of prefixes to sample from in each iteration}. Each figure shows a representative frame from each video and experiment configuration. The examples in row one use the texts: $T_{Global}$:``Origami swan with white paper skin,'' $T_{Local}$: ``Origami white paper skin,'' and in row two: $T_{Global}$:``Dog with bengal tiger fur,'' $T_{Local}$: ``Bengal tiger fur.'' (a) no prefixes, (b) $4$ local, no global, (c) $4$ global, no local (d) $4$ global \& $4$ local, (e) $8$ global \& $8$ local. The specific prefixes are described in~\cref{appendix_text_augmentations}.}
    \label{fig:prefix_exps}
    \vspace{-1.0cm}
\end{figure}

\begin{figure}
\centering
\begin{tabular}{ccccc}
         \includegraphics[width=0.191\linewidth]{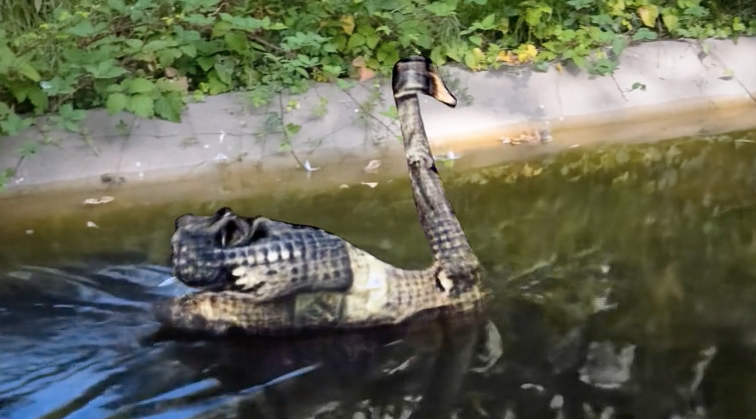} &
         \includegraphics[width=0.191\linewidth]{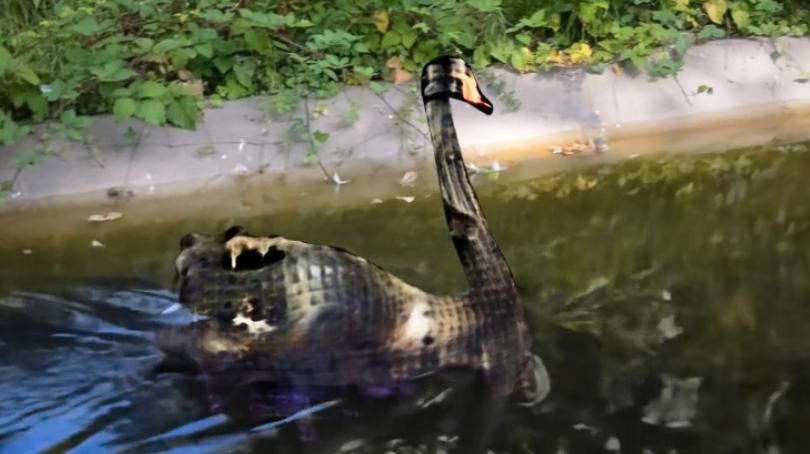}&
         \includegraphics[width=0.191\linewidth]{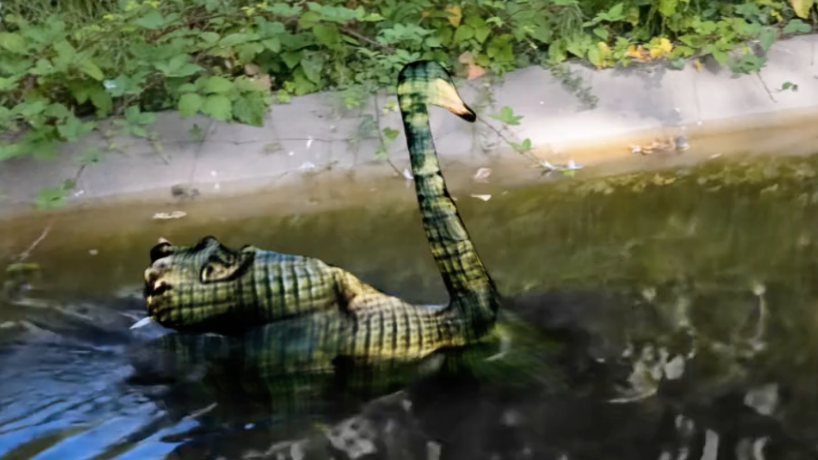} &
         \includegraphics[width=0.191\linewidth]{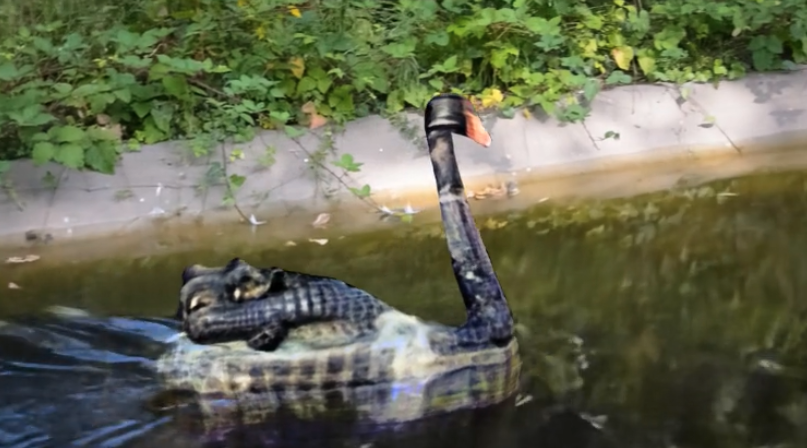} &
         \includegraphics[width=0.191\linewidth]{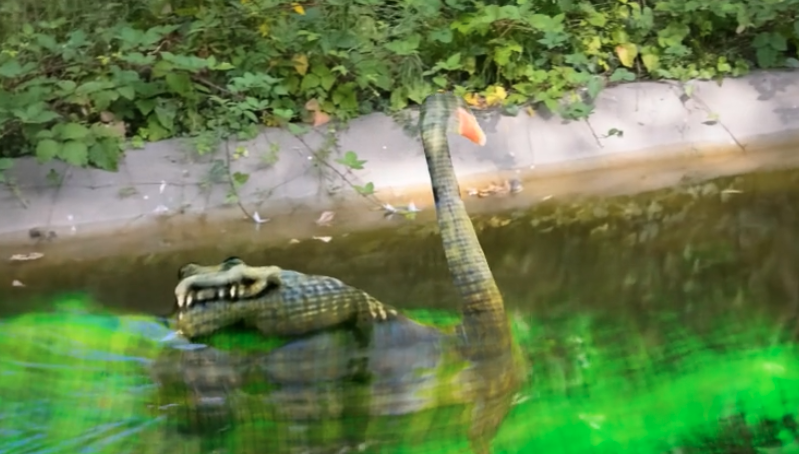} 
         \\ (a) & (b) & (c) & (d) & (e)
\end{tabular}
     \caption{\textbf{Ablation on each loss term} - All experiments were run with the same seed and with the texts: $T_{Global}$: ``A swan with crocodile skin,'' $T_{Local}$: ``Crocodile skin.'' 
     Each figure shows a representative frame from a video, where we ablate one of our loss terms in Eq.~\ref{objective function}. (a) All losses, (b) w/o local loss, (c) w/o global loss, (d) w/o temporal loss, (e) w/o sparsity loss.\label{fig:my_ablation_studylabel_croc}} 
     \vspace{-0.5cm}
\end{figure}

\begin{figure}
\centering
\begin{tabular}{ccc}
        
         \includegraphics[width=0.33\linewidth]{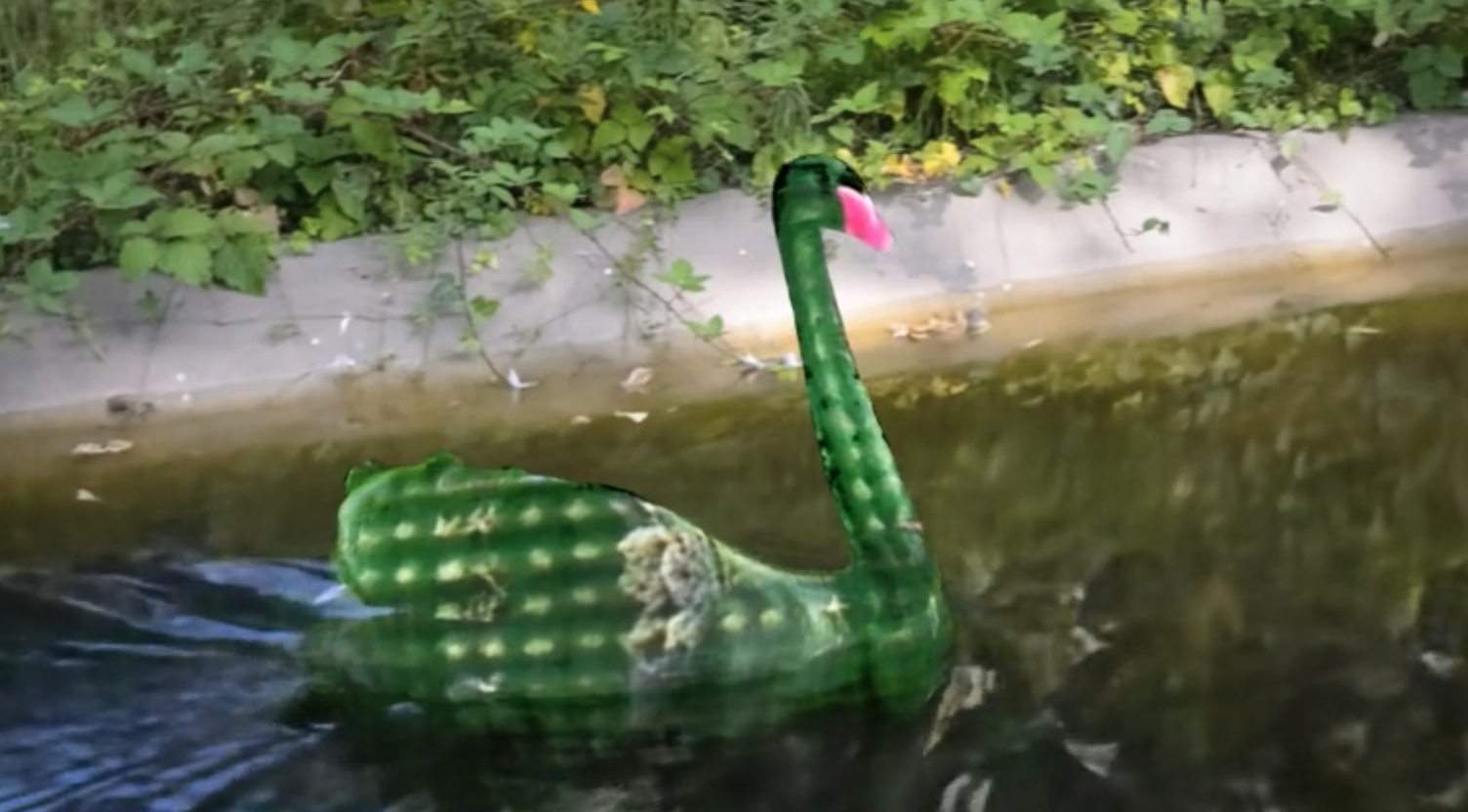} &
         \includegraphics[width=0.33\linewidth]{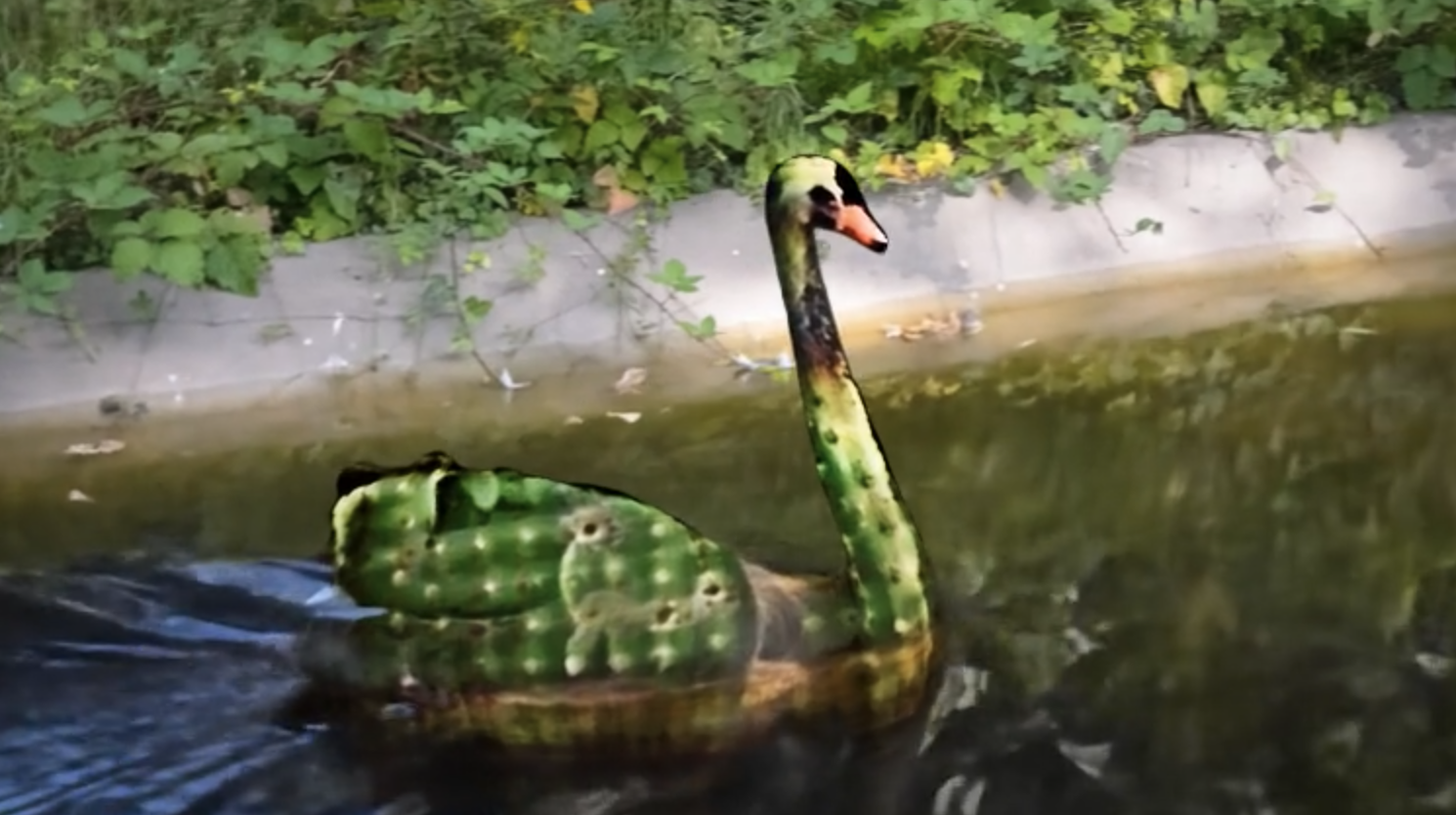} &
         \includegraphics[width=0.33\linewidth]{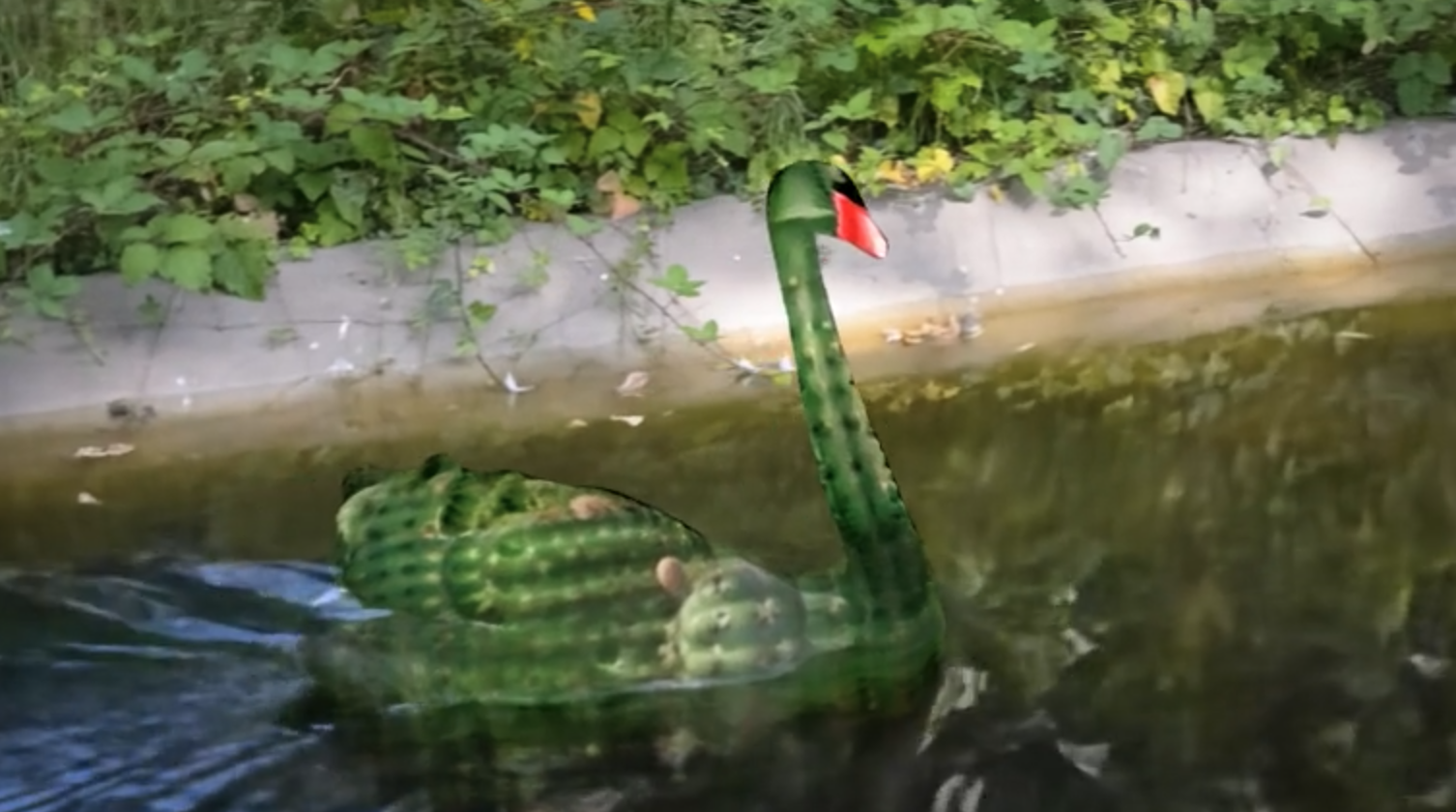} \\
         (a) & (b) & (c)
\end{tabular}
     \caption{\textbf{Global and local semantics}. A representative frame from each of the edited video frames. The texts used are: (a) $T_{Global}$: ``Swan with cactus skin,'' $T_{Local}$: ``Cactus skin,'' (b) $T_{Global}$: ``Swan with cactus skin,'' $T_{Local}$: ``Rough cactus skin,'' (c) $T_{Global}$: ``Swan made out of cactus,'' $T_{Local}$: ``Catus skin.'' (a) and (b) have the same global target text while (a) and (c) have the same local target texts. In (b) the local details have changed to a more rough texture. In (c) the global swan semantics are better preserved.} 
    \label{fig:Ablation_local_global}
\end{figure}

\subsection{Text augmentation} \label{section:text_aug}

We add textual augmentation to our method to address some of the challenges with prompt engineering. Inspired by Zhou et al.~\cite{LearningToPrompt}, we add neutral prefixes to both the local and global target texts, e.g., ``a photo of a \{\}.'' We then sample a new prefix each iteration for each of the target texts as a form of regularization.
Fig.~\ref{fig:prefix_exps} illustrates our text augmentation experiment. We demonstrate that using prefixes increases the robustness and quality of the results. In each experiment and each iteration, we sample one prefix among a set of prefixes (details in appendix ~\cref{appendix_text_augmentations})
for both the global and local target texts. In this experiment, we vary the number of prefixes to sample from and show that using an equal amount of prefixes for both the local and global target texts produces better quality results than using no prefixes. In both the swan and dog examples for columns (e) and (d) that use multiple prefixes, we see that the stylizations are more detailed than the examples in columns (a-c), e.g., in the dog example (d) and (e) the tiger fur has white pigments around the belly area and also more natural tiger stripes.

\subsection{Ablation study}\label{section:ablation_study}
We validate each term in our objective function (Eq.~\ref{objective function}) with an ablation study illustrated in Fig.~\ref{fig:my_ablation_studylabel_croc}.
In (a), we see that all losses combined result in a natural stylization that shows clear characteristics of a swan with crocodile skin. In (b) we see that without the \textbf{local loss} the results lack fine-grain details. It is still evident what the intended stylization was but the crocodile texture is not as clear as in (a). The results in (c) show that without the \textbf{global loss} the global semantics are less clear, e.g., the swan's neck is not stylized as detailed as the rest of the body.
In (d), we see that without the \textbf{temporal loss}, we get more edits outside the mask of the object. In (e), we see that without the \textbf{sparsity loss}, the stylization is very noisy.

\begin{figure}
\centering
\begin{tabular}{ccc}
        
         \includegraphics[width=0.33\linewidth]{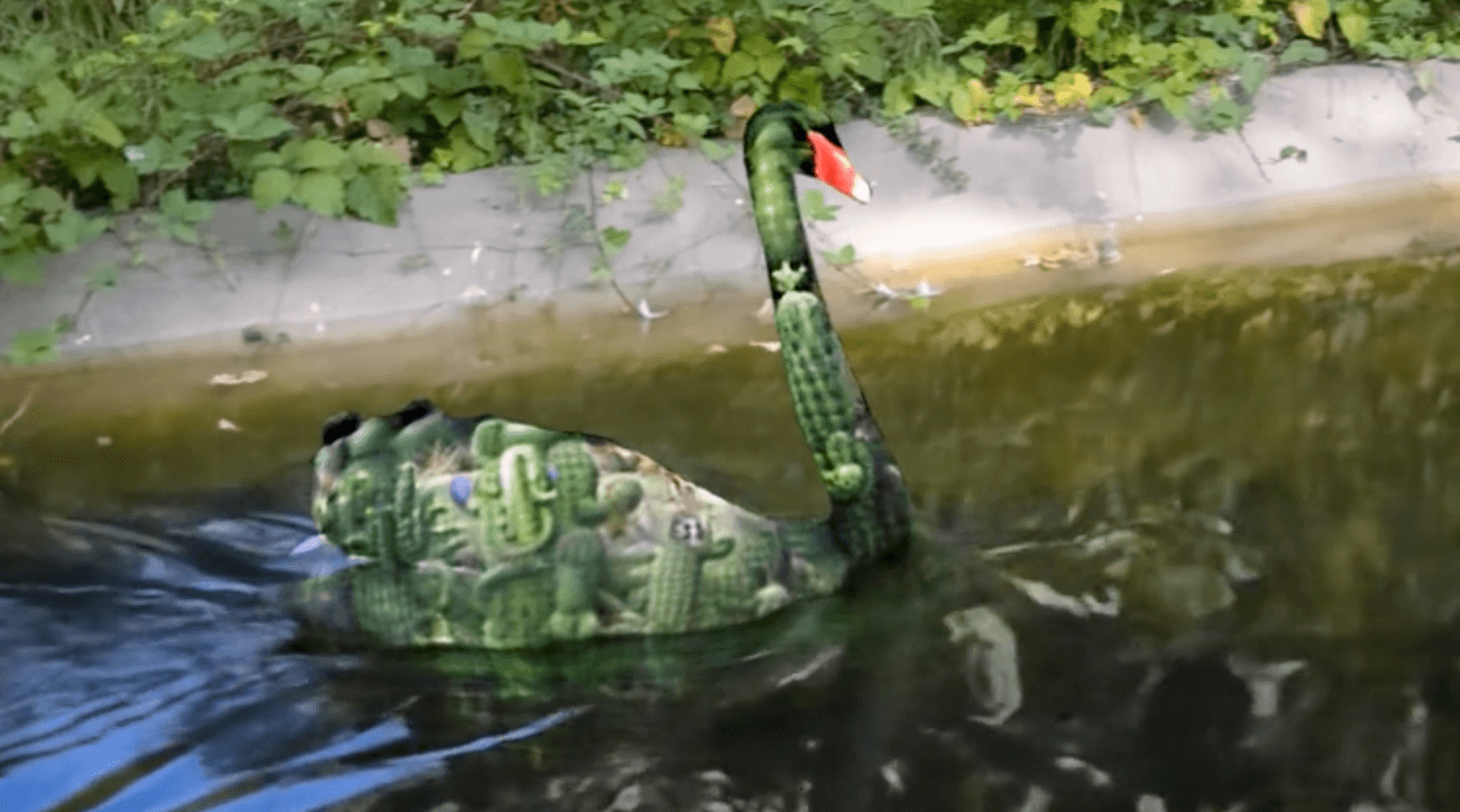} &
         \includegraphics[width=0.33\linewidth]{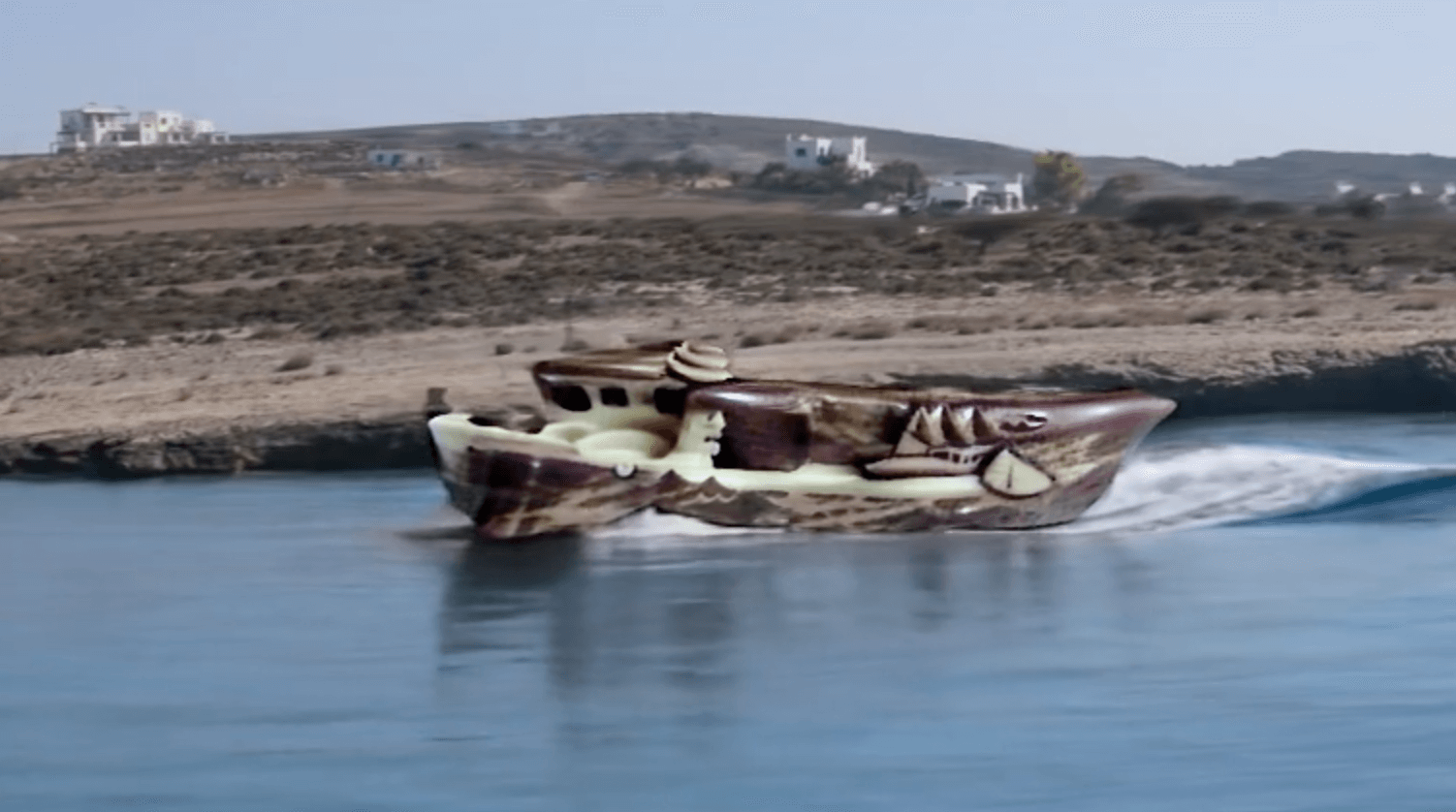} &
         \includegraphics[width=0.33\linewidth]{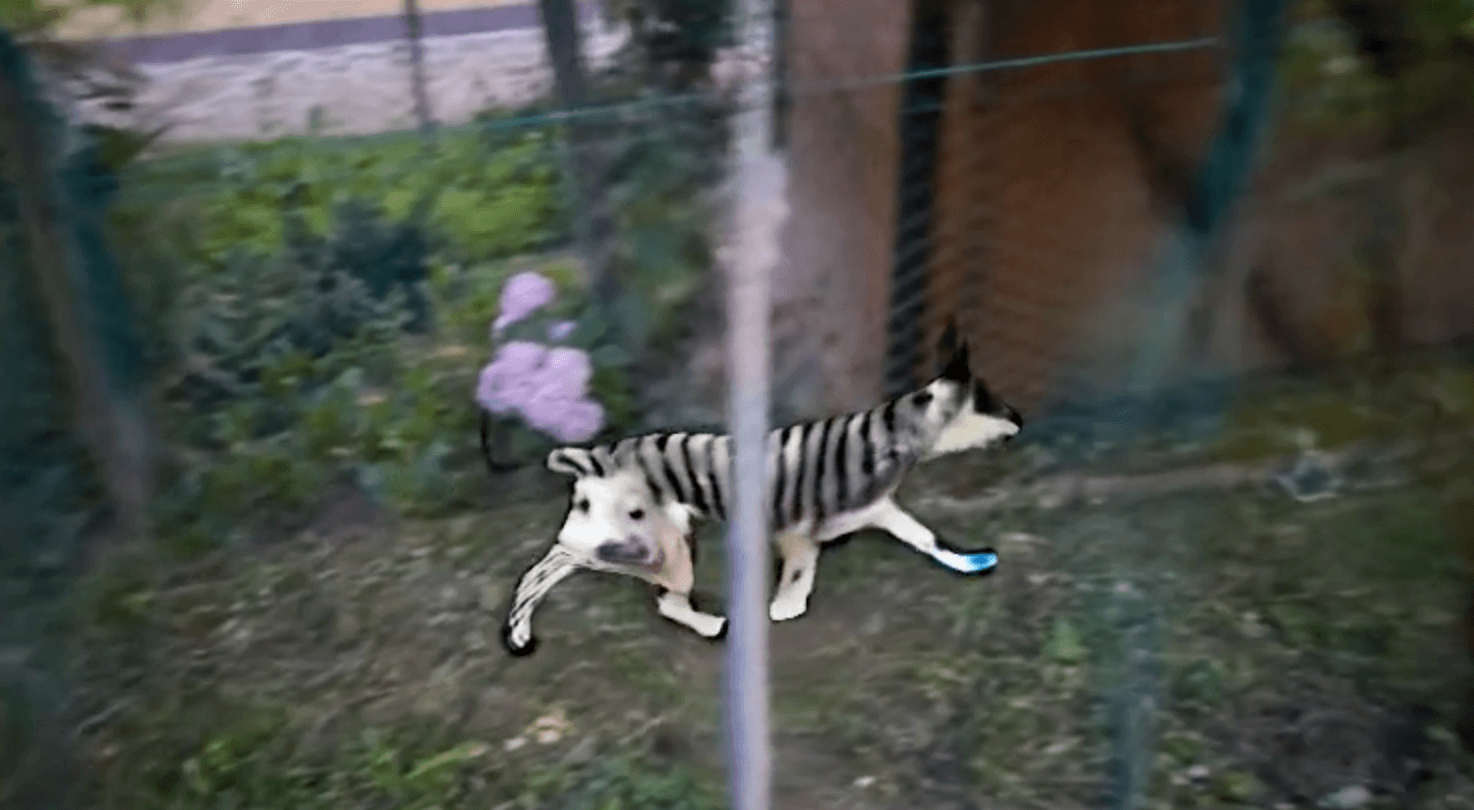}  \\
         (a) & (b) & (c) 
\end{tabular}
     \caption{\textbf{Limitations}. A representative frame from each of the edited video frames. The texts used are: (a) $T_{Global}$: ``Swan with cactus skin,'' $T_{Local}$: ``Cactus skin,'' (b) $T_{Global}$: ``Boat made out of chocolate,'' $T_{Local}$: ``Chocolate texture,'' (c) $T_{Global}$: ``Dog with zebra fur,'' $T_{Local}$: ``Zebra fur.'' In (a), the cactus texture is applied like photos of cactus. In (b) the chocolate boat has a sailing ship printed at the end of the boat. In (c), the dog has a face of a dog on its haunches.} 
    \label{fig:limitations}
\end{figure}

In Fig~\ref{fig:Ablation_local_global} we illustrate how the local loss affects the fine-grained details and the global loss affects the global semantics. We use (a) as a baseline and then vary the local target texts in (b) and the global target texts in (c). In (b) we see, how changing the local target text to include ``rough'' affects the details of the cactus texture. In (c) we see how the global semantics of the swan's body become more realistic as an effect of changing the global target text.

\subsection{Limitations} \label{section:limitations}
Fig.~\ref{fig:limitations} illustrates some of the limitations of our method. During training our method occasionally starts to overfit and produce unintended stylizations. In (a), we see that the body of the swan contains photo-like cacti instead of natural cactus texture. In (b) and (c), we see how our model has used the contexts of the global target text. In (b), a sailing ship has been added to the end of the boat and in (c), a face of a dog has been added. 

Our method is limited to text-prompts that do not entail features that cross between the decomposed atlas layers. Some changes are best realized through a shape change, e.g, ``Swan with long hair.'', which is not currently possible in our framework.
Other limitations of our method include stylizations that focus too much on some property of the target texts, e.g., we experienced that the global target texts and similar variants ``Swan with strawberry texture skin,'' gave noisy results, where the swan was colored red.

\section{Conclusion}

We considered the problem of developing intuitive and semantic control for consistent editing and styling of objects in videos. This problem poses a challenge in generating consistent content and style changes over time while being able to produce fine-grained details and preserve the global semantics. We proposed a method that uses CLIP~\cite{CLIP} and the video object representation~\cite{LayeredNeuralAtlases} to stylize objects in videos by using both a global target text, to control the global semantics of the stylization, and a local target text, to control the fine-grained details. We demonstrated that the specificity and the prefixes of the target texts can have a significant impact on the details produced by our method's stylization. In future work, it would be interesting to investigate the limitations of CLIP. A model that is able to generate fine-grained stylizations of videos and images could be leveraged to create data augmentations in other learning settings.
Another line of future work is to extend our model to be able to generate shape changes or even new objects from scratch.

\subsection*{Acknowledgement}
This research was supported by the Pioneer Centre for AI, DNRF grant number
P1. We would like to thank Ira Assent for the helpful discussions.

\clearpage
%
%
\bibliographystyle{splncs04}

\section{Appendix}
\subsection{Training details}
We train on a single GPU (RTX $6000$). Our method is implemented using the PyTorch framework \cite{PyTorch} and will be made available. We use an ADAM optimizer \cite{ADAM} with an initial learning rate of $1^{-4}$ and decay the learning rate by a factor of $0.9$ every $200$ iterations. Our method takes about $40$ minutes and $2000$ iterations with a batch size of three input frames, each of size $432$x$768$, sampled from a video of maximally $70$ frames. High-quality results usually appear after $10-20$ minutes.

All images are normalized and resized to $224$x$224$ before being passed to the CLIP model. We normalization with mean $(0.48145466, 0.4578275, 0.40821073)$ and standard deviation $(0.26862954, 0.26130258, 0.27577711)$, which is the same as CLIP was trained with. We use the pretrained ``ViT-L/14'' CLIP model loaded from OpenAi's \href{https://github.com/openai/CLIP}{GitHub page}. 
\subsection{Image augmentation}
\label{sec:appendix_img_augmentations}
\begin{itemize}
    \item \textit{Random Crops: }For the global views we crop the reconstructed frame with a random scaling in the range:  $[0.9,1.0]$ and for the local views we use a scaling in the range $[0.1,0.5]$. For both global and local views we use a random aspect ratio in the range $[0.8,1.2]$. The local crops have a chance of not including the object we want to edit, e.g. a small crop from the bounding box containing the Swan in Fig.~\ref{fig:varity_examples} has a chance of only containing the background. To combat this, we use the $\alpha$-map to locate the object, and sample local crops until we get a crop that contains at least $\frac{1}{3}$ of the object.
    \item \textit{Random Perspective Transformation:} With probability $\frac{1}{2}$, we sample a number $d$ uniformly at random in the range: $[0.1, 0.5]$ and use $d$ as the distortion scale.
    \item \textit{Random Background Removal:} We locate the foreground object using the $\alpha$-map. Then, with probability $\frac{1}{2}$ we color all background pixels black. 
\end{itemize}

\subsection{Text augmentations}
\label{appendix_text_augmentations}

In our text augmentation we randomly sample one of the following prefixes in each iteration:
\begin{enumerate}
    \item ``a photo of a \{\}''
    \item ``a \{\}''
    \item ``an image of a \{\}''
    \item ``the \{\}''
    \item ``image of a \{\}''
    \item ``image of the \{\}''
    \item ``photo of a \{\}''
    \item ``photo of the \{\}''
\end{enumerate}
All prefixes are neutral and intended to not change the semantics of the stylization. In the text augmentation experiments (Sec. \ref{section:text_aug}), if we use four prefixes, these are the first four in the list above.

\end{document}